\documentclass[]{article}

\usepackage{amsmath}

\usepackage{amsthm}
\usepackage{amssymb}
\usepackage{amsfonts}
\usepackage{enumitem}
\usepackage{tikz}
\usetikzlibrary{shapes,arrows}

\newcommand{\sys}[1]{\ensuremath{\mathbf{#1}}}
\newcommand{\Cn}[3]{\ensuremath{\mathrm{Cn}_{\mathbf{#1}}^{#2}(#3)}}

\newcommand{\verberg}[1]{}

\newcommand{\SSs}{\ensuremath{\mathsf{S}}}
\newcommand{\UUs}{\ensuremath{\mathsf{U}}}
\newcommand{\OOs}{\ensuremath{\mathsf{O}}}

\newcommand{\Cs}{\ensuremath{\mathsf{C}}}

\newcommand{\Ra}{\ensuremath{\Rightarrow}}

\newcommand{\nc}{\,\mid\!\sim\,}

\tikzstyle{hlred}=[fill=red]
\tikzstyle{hlin}=[fill=green!20, draw=green!50]
\tikzstyle{hlout}=[fill=red!20, draw=red!50]
\tikzstyle{hlund}=[fill=black!20, draw=black!50]
\tikzstyle{hlwhi}=[fill=white, draw=black]
\usetikzlibrary{snakes,arrows,shapes}

\newcounter{Gcount}
\newcommand{\Gtel}{\refstepcounter{Gcount}\theGcount}

\title{A structured argumentation framework for detaching conditional obligations}
\author{Mathieu Beirlaen and Christian Stra{\ss}er \\ \small Research Group for Non-Monotonic Logics and Formal Argumentation, \\ \small Institute for Philosophy II, Ruhr University Bochum \\ \small {Email: mathieu.beirlaen@rub.de, christian.strasser@rub.de.}\footnote{The research of both authors was supported by a Sofja Kovalevskaja award of the Alexander von Humboldt-Foundation, funded by the German Ministry for Education and Research.}}

\begin{document}

\theoremstyle{definition1}
\newtheorem{definition1}{Definition}
\newtheorem{example1}{Example}
\newtheorem{fact1}{Fact}
\newtheorem{property}{Property}
\newtheorem{lemma1}{Lemma}
\newtheorem{theorem1}{Theorem}

\maketitle

\begin{abstract}
We present a general formal argumentation system for dealing with the detachment of conditional obligations. Given a set of facts, constraints, and conditional obligations, we answer the question whether an unconditional obligation is detachable by considering reasons for and against its detachment. For the evaluation of arguments in favor of detaching obligations we use a Dung-style argumentation-theoretical semantics. We illustrate the modularity of the general framework by considering some extensions, and we compare the framework to some related approaches from the literature.
\end{abstract}

  \begin{center}
  \end{center}

\section{Introduction}

We take an argumentative perspective on the problem of detaching conditional obligations relative to a set of facts and constraints. We allow for the construction of arguments the deontic conclusions of which are candidates for detachment. Next, we define a number of ways in which these arguments may attack one another, as when the conclusions of two arguments are conflicting. We borrow Dung's semantics \cite{Dun95} for evaluating arguments relative to the attack relations that hold between them. Conclusions of arguments which are evaluated positively are safely detachable in our framework. They can be interpreted as all-things-considered obligations -- following Ross \cite{Ros30} -- or output obligations -- following Makinson \& van der Torre \cite{MakTor00,MakTor01}.

The argumentative approach defended in this paper is both natural and precise. Norms which guide reasoning are naturally construed as conclusions of proof sequences. Objections raised against the derivation of certain obligations are naturally construed as argumentative attacks. Arguments are naturally evaluated in terms of the objections raised against them.

In Section \ref{sec_basicframe} we introduce a basic argumentation system for evaluating arguments the conclusions of which can be interpreted as all-things-considered obligations. This generic, modular framework can be extended in various ways, as we illustrate in Section \ref{sec_extensions}. We show how various mechanisms for conflict-resolution can be implemented (Section \ref{sub_confl}), and how we can rule out obligations committing us to further violations or conflicts (Section \ref{sub_shadow}). In Section \ref{sec_related} we compare our approach to related systems from the literature. We end by pointing to some further expansions of our framework, which we aim to present in a follow-up paper (Section \ref{sec_outlook}).

%
%
%
%

\section{The basic framework}\label{sec_basicframe}

We start by reviewing the basic concepts needed from Dung's semantics (Section \ref{sub_abstr}). Next we turn to the construction of deontic arguments (Section \ref{sub_darg}) and attack definitions (Section \ref{sub_attack}). We define a consequence relation for detaching all-things-considered obligations in deontic argumentation frameworks (Section \ref{sub_deval}), and present some of its meta-theoretical properties (Section \ref{sub_meta}).

\subsection{Abstract argumentation}\label{sub_abstr}

A Dung-style \emph{abstract argumentation framework} (AF) is a pair $(\mathcal{A},\mathsf{Att})$ where $\mathcal{A}$ is a set of arguments and $\mathsf{Att} \subseteq \mathcal{A}\times\mathcal{A}$ is a binary relation of attack. Relative to an AF, Dung defines a number of extensions -- subsets of $\mathcal{A}$ -- on the basis of which we can evaluate the arguments in $\mathcal{A}$.
\begin{definition1}[Complete and grounded extension]\label{def_af}
  Let $(\mathcal{A}, {\sf Att})$ be an AF. For any $a \in \mathcal{A}$, $a$ is acceptable w.r.t.\ some $\mathcal{S} \subseteq \mathcal{A}$  (or, $\mathcal{S}$ defends $a$) iff for all $b$ such that $(b,a) \in {\sf Att}$ there is a $c \in \mathcal{S}$ for which $(c,b) \in {\sf Att}$.\\
If $\mathcal{S} \subseteq \mathcal{A}$ is conflict-free, i.e.\ there are no $a,b\in \mathcal{S}$ for which $(a,b)\in {\sf Att}$, then:
\begin{itemize}[itemsep=-1mm]
\item $\mathcal{S}$ is a complete extension iff $a \in \mathcal{S}$ whenever $a$ is acceptable w.r.t.\ $\mathcal{S}$;
\item $\mathcal{S}$ is the grounded extension iff it is the set inclusion minimal complete extension.
\end{itemize}
\end{definition1}
\noindent Dung \cite{Dun95} showed that for every AF there is a grounded extension, it is \emph{unique}, and it can be constructed as follows.

\begin{definition1}[Defense]
A set of arguments $\mathcal{X}$ defends an argument $a$ iff every attacker of $a$ is attacked by some $b \in\mathcal{X}$.
\end{definition1}
\begin{definition1}[Construction of the grounded extension]\label{def_ground2}
The grounded extension $\mathcal{G}$ relative to an AF $(\mathcal{A}, {\sf Att})$ is defined as follows (where $\mathcal{A}$ is countable):
\begin{itemize}
\item $\mathcal{G}_0$: the set of all arguments in $\mathcal{A}$ without attackers;
\item $\mathcal{G}_{i+1}$: all arguments defended by $\mathcal{G}_i$;
\item $\mathcal{G} = \bigcup_{i\geq 0} \mathcal{G}_i$
\end{itemize}
\end{definition1}
Besides the grounded extension, a number of further extensions (preferred, (semi-)stable, ideal etc.) have been defined in the literature. Due to space limitations, we focus exclusively on grounded extensions in the remainder.

On Dung's abstract approach \cite{Dun95}, arguments are basic units of analysis the internal structure of which is not represented. But nothing prevents us from \emph{instantiating} such abstract arguments by conceptualizing them as proof trees for deriving a conclusion based on a set of premises and inference rules. Frameworks with instantiated arguments are called \emph{structured argumentation frameworks} (for examples, see e.g.\ \cite{StrArg14}).\footnote{Our approach is similar in spirit to the $ASPIC^+$ framework for structured argumentation from e.g.\ \cite{ModPra14}. We return to this point in Section \ref{sub_argfranor}.} In the remainder of Section \ref{sec_basicframe} we show how questions regarding obligation detachment in deontic logic can be addressed and answered within structured \emph{deontic} argumentation frameworks.

\subsection{Instantiating deontic arguments}\label{sub_darg}

Our formal language $\mathcal{L}$ is defined as follows:
\begin{tabbing}
$\mathcal{P}$\quad \= $:=$ \= $\{p,q,r,\ldots\}$\quad\quad\quad\quad\quad\quad\quad\quad\quad\quad\quad\quad\quad \= $\mathcal{L}^{\Ra}$ \= $:=$ \= $\langle\mathcal{L}^P\rangle\Ra\langle\mathcal{L}^P\rangle$\\
$\mathcal{L}^P$ \> $:=$ \> $\mathcal{P}\mid \top \mid \bot \mid \neg\langle\mathcal{L}^P\rangle \mid \langle\mathcal{L}^P\rangle\vee\langle\mathcal{L}^P\rangle$ \>
$\mathcal{L}^{\OOs}$ \> $:=$ \> $\OOs\langle\mathcal{L}^P\rangle$\\
$\mathcal{L}^{\Box}$ \> $:=$ \> $\Box\langle\mathcal{L}^P\rangle\mid\langle\mathcal{L}^P\rangle\mid \neg\langle\mathcal{L}^{\Box}\rangle\mid \langle\mathcal{L}^{\Box}\rangle\vee\langle\mathcal{L}^{\Box}\rangle$ \>
$\mathcal{L}$ \> $:=$ \> $\mathcal{L}^P \mid \mathcal{L}^{\Box} \mid \mathcal{L}^{\Ra} \mid \mathcal{L}^{\OOs}$\\
\end{tabbing}
\vspace{-15pt}
The classical connectives $\wedge,\supset,\equiv$ are defined in terms of $\neg$ and $\vee$.
We represent \emph{facts} as members of $\mathcal{L}^P$. Where $A,B\in\mathcal{L}^P$, \emph{conditional obligations} are formulas of the form $A\Ra B$, read `If $A$, then it -- prima facie -- ought to be that $B$' or `If $A$, then $B$ is prima facie obligatory'.\footnote{Depending on the context of application, the following alternative readings are also fine: `If $A$ is the case, then $B$ is pro tanto obligatory', `If $A$, then the agent ought (prima facie, pro tanto) to bring about $B$'. On the latter, agentive reading, we can think of `\Ra' as implicitly indexed by an agent.} Where $A\in\mathcal{L}^P$, a \emph{constraint} $\Box A$ abbreviates that $A$ is settled, i.e.\ that $A$ holds unalterably.\footnote{\label{fn_unalterable}If $\Box A$ holds, then the fact that $A$ is deemed fixed, necessary, and unalterable. Obligations which contradict these facts are unalterably violated. Carmo \& Jones cite three factors giving rise to such unalterable violations. The first is time, e.g.\ when you did not return a book you ought to have returned by its due date. The second is causal necessity, e.g.\ when you killed a person you ought not to have killed. The third is practical impossibility, e.g.\ when a dog owner stubbornly refuses to keep her dog against the house regulations, and nobody else dares to try and convince her to remove it \cite[pp.~283-284]{CarJon02}.} Formulas of the form $\OOs A$ (where $A\in\mathcal{L}^P$) represent \emph{all-things-considered} obligations.

Unless specified otherwise, upper case letters $A,B,\ldots$ denote members of $\mathcal{L}^P$ and upper case Greek letters $\Gamma,\Delta,\ldots$ denote subsets of $\mathcal{L}^P \cup \mathcal{L}^{\Box} \cup \mathcal{L}^{\Ra}$. Where $\Gamma\subseteq\mathcal{L}$ and $\dag\in\{P,\Box,\Ra,\OOs\}$, $\Gamma^{\dag} = \Gamma\cap\mathcal{L}^{\dag}$.

$\Cn{CL}{}{\Gamma}$ denotes the closure of $\Gamma\subseteq\mathcal{L}^{P}$ under propositional classical logic, \sys{CL}. $\Cn{L^{\Box}}{}{\Gamma}$ denotes the closure of $\Gamma\subseteq\mathcal{L}^{\Box}$ under \sys{L^{\Box}}, which we use as a generic name for a modal logic for representing background constraints, e.g.\ \sys{T}, \sys{S4}, \sys{S5}, etc. In our examples below, we will assume that \sys{L^{\Box}} is normal and validates the axiom $\Box A\supset A$.\footnote{\label{fn_BoxRa}Moreover, where $\Delta^{\Ra}\subseteq\mathcal{L}^{\Ra}$, we assume that $\Gamma\vdash_{\sys{L^{\Box}}}\Box A$ iff $\Gamma\cup\Delta^{\Ra}\vdash_{\sys{L^{\Box}}}\Box A$.}

\emph{Arguments} are ordered pairs $\langle A: {\sf s}\rangle$ in which $A$ is called the \emph{conclusion}, and ${\sf s}$ a \emph{proof sequence} for deriving $A$. We use lower case letters $a,b,c,\ldots$ as placeholders for arguments.

\begin{definition1}\label{def_arg}
Given a premise set $\Gamma$, we allow the following rules for constructing arguments:
\begin{itemize}
\item[(i)] If $\Box A\in\Cn{L^{\Box}}{}{\Gamma}$, then $\langle \Box A: -- \rangle$ is an argument; (where $--$ denotes the empty proof sequence)
\item[(ii)] If $A\Ra B \in \Gamma^{\Ra}$ and $A\in\Cn{L^{\Box}}{}{\Gamma}$, then $\langle \OOs B: A, A\Ra B \rangle$ is an argument;
\item[(iii)] If $A\Ra B \in \Gamma^{\Ra}$ and $a = \langle \OOs A: \ldots \rangle$ is an argument, then $\langle \OOs B: a, A\Ra B \rangle$ is an argument;
\item[(iv)] If $a = \langle \OOs A: \ldots \rangle$ and $b = \langle \OOs B: \ldots \rangle$ are arguments, then $\langle \OOs(A\wedge B) : a,b\rangle$ is an argument.
\item[(v)] If $a = \langle \OOs A: \ldots \rangle$ is an argument and $\Box(A \supset B) \in \Cn{L^{\Box}}{}{\Gamma}$, then $\langle \OOs B: a, \Box(A\supset B) \rangle$ is an argument.
\end{itemize}
Argument $a$ is a \emph{deontic argument} if $a$ is of the form $\langle \OOs A: \ldots \rangle$.
We use $\Cs(a)$ to denote the set of all formulas in $\mathcal{L}$ used in the construction of $a$, including its conclusion. E.g.\ where $a=\langle \OOs q: p, p\Ra q \rangle$ and $b=\langle \OOs r: a, q\Ra r \rangle$, $\Cs(a) = \{p, p\Ra q, \OOs q\}$ and $\Cs(b)= \{p, p\Ra q, \OOs q, q\Ra r, \OOs r\}$.
Argument $a$ is a \emph{sub-argument} of argument $b$ if $\Cs(a)\subseteq\Cs(b)$; $a$ is a \emph{proper} sub-argument of argument $b$ if $\Cs(a)\subset\Cs(b)$; and $b$ is a \emph{super-argument} of argument $a$ if $a$ is a proper sub-argument of $b$.
\end{definition1}
(ii)-(v) correspond to inference rules well-known from deontic logic. (ii) allows for the factual detachment of an all-things-considered obligation $\OOs B$ from a conditional prima facie obligation $A\Ra B$ and a fact $A$. (iii) is a deontic detachment principle. (iv) and (v) allow for obligation aggregation (or agglomeration), resp.\ inheritance (or weakening).

\begin{example1}[Constructing arguments]\label{ex_ctd}
Let $\Gamma_{\Gtel\label{gctd}} = \{\Box p, \top\Ra\neg p, \neg p\Ra\neg q, p\Ra q\}$. By Definition \ref{def_arg} we can construct -- amongst others -- the following arguments from $\Gamma_{\ref{gctd}}$:\smallskip

\begin{tabular}{l l l l}
$a_1$: & $\langle \Box p: -- \rangle$ & \quad$a_4$: & $\langle \OOs q: p, p\Ra q\rangle$ \\
$a_2$: & $\langle \OOs\neg p: \top, \top\Ra\neg p \rangle$ & \quad$a_5$: & $\langle\OOs (\neg q\wedge q): a_3,a_4\rangle$ \\
$a_3$: & $\langle \OOs \neg q: a_2, \neg p\Ra\neg q\rangle$ & \quad$a_6$: & $\langle \OOs(q\vee r): a_4, \Box(q\supset(q\vee r)) \rangle$ \\
\end{tabular}\smallskip

Argument $a_1$ is constructed from $\Box p\in\Gamma_{\ref{gctd}}$ in view of (i). Arguments $a_2$ and $a_4$ are constructed by means of (ii)\footnote{Note that, in the construction of argument $a_4$, the formula $p$ follows from $\Gamma_{\ref{gctd}}$ by $\Box p$ and since $\vdash_{\bf L^\Box} \Box p \supset p$.}; $a_3$ is constructed from $a_2$ by means of (iii); $a_5$ is constructed from $a_3$ and $a_4$ by (iv); and $a_6$ is constructed from $a_4$ by (v).
\end{example1}
We can interpret $\Gamma_{\ref{gctd}}$ as representing a classic \emph{contrary-to-duty} (CTD) scenario (for the sake of readability, we omit the qualifier `prima facie' in our reading of conditional obligations):\footnote{The example is adapted from \cite{PraSer96}.}\smallskip

\noindent\begin{tabular}{l p{9.8cm}}
$\top\Ra\neg p$ & There ought not be a dog.\\
$\neg p\Ra\neg q$ & If there is no dog, there ought not be a warning sign.\\
$p\Ra q$ & If there is a dog, there ought to be a warning sign.\\
$\Box p$ & It is settled that there is a dog.\\
\end{tabular}\smallskip

Of course, not all of the conclusions of arguments $a_2$-$a_6$ qualify as all-things-considered obligations. Argument $a_5$, for instance, is internally incoherent and should be filtered out when evaluating the arguments constructed from $\Gamma_{\ref{gctd}}$. Arguments are evaluated in terms of the attack relations which hold amongst them.
Before we turn to the definition of these relations, we point out that rules (i)-(v) in Definition \ref{def_arg} allow for a version of the necessitation rule whenever \sys{L^{\Box}} is a normal modal logic. For instance, given a premise set $\{\Box p, \top\Ra q\}$, we can construct the argument $a_1=\langle\OOs q: \top, \top\Ra q \rangle$ by (ii). Since $\Box p\vdash_{\sys{L^{\Box}}}\Box(q\supset p)$, we can construct the argument $a_2=\langle \OOs p: a_1, \Box(q\supset p) \rangle$ by (v). If desired, the construction of $a_2$ can be prevented by defining -- in addition to `$\supset$' -- a weaker (non-material) implication connective in \sys{L^{\Box}} on the basis of which to construct arguments in line with clause (v) in Definition \ref{def_arg}.

\subsection{Attacking deontic arguments}\label{sub_attack}

In our basic framework, we define two ways in which arguments may attack one another. First, we take care that unalterably violated obligations are attacked by the constraints which violate them. (We write $A = {-} B$ in case $A = \neg B$ or $B = \neg A$.)
\begin{definition1}[Fact attack]\label{def_fattack}
Where $a = \langle \OOs A : \ldots \rangle$ is an argument, let $\UUs\OOs(a) = \{B\mid \OOs B\in \Cs(a)\}$. Where $\emptyset\neq\Theta\subseteq\UUs\OOs(a)$, $\langle \Box{-}\bigwedge\Theta: -- \rangle$ attacks $a$.
\end{definition1}
In Example \ref{ex_ctd} the obligation $\OOs\neg p$ cannot guide the agent's actions, since it cannot be acted upon in view of the constraint $\Box p$. Definition \ref{def_fattack} takes care that $a_1$ attacks $a_2$, since $\UUs\OOs(b) = \{\neg p\}$. Note that, as soon as $A\in\UUs\OOs(a)$ for some argument $a$ and formula $A$, $A\in\UUs\OOs(b)$ for any super-argument $b$ of $a$. Consequently, if an argument $c$ attacks $a$ in view of Definition \ref{def_fattack}, then $c$ also attacks all super-arguments $b$ of $a$. So in Example \ref{ex_ctd} the argument $a_1$ attacks $a_2$ as well as its super-arguments $a_3$ and $a_5$.

Since we assume that \sys{L^{\Box}} is a normal modal logic, we know that $\Box(\neg(\neg q\wedge q))\in\Cn{L^{\Box}}{}{\Gamma_{\ref{gctd}}}$. Hence, by Definition \ref{def_fattack} again, argument $a_7 = \langle \Box(\neg(\neg q\wedge q)): -- \rangle$ attacks argument $a_5$ from Example \ref{ex_ctd}.
\begin{example1}[Attacks on incoherent arguments]\label{ex_incoh}
Let $\Gamma_{\Gtel\label{ginc}} = \{\top\Ra p, \top\Ra \neg p, \top\Ra q\}$. We construct the following arguments on the basis of $\Gamma_{\ref{ginc}}$:\smallskip

\noindent\begin{tabular}{l l l l}
$a_1$: & $\langle \OOs p: \top, \top\Ra p\rangle$ & \quad$a_4$: & $\langle \OOs(p\vee\neg q): a_1, \Box(p\supset(p\vee\neg q))\rangle$ \\
$a_2$: & $\langle \OOs\neg p: \top, \top\Ra\neg p\rangle$ & \quad$a_5$: & $\langle \OOs(\neg p\wedge (p\vee\neg q): a_2, a_4\rangle$ \\
$a_3$: & $\langle \OOs q: \top, \top\Ra q\rangle$ & \quad$a_6$: & $\langle\OOs \neg q: a_5, \Box((\neg p\wedge(p\vee\neg q))\supset \neg q)\rangle$ \\
\end{tabular}\smallskip

\noindent By Definition \ref{def_fattack}:\smallskip

\noindent $\UUs\OOs(a_5) = \{p, p\vee\neg q, \neg p, \neg p\wedge(p\vee\neg q)\}$\\
$\UUs\OOs(a_6) = \{p, p\vee\neg q, \neg p, \neg p\wedge(p\vee\neg q), \neg q\}$\smallskip

\noindent Hence, both $a_5$ and $a_6$ are attacked by $a_7$:\smallskip

\noindent$a_7 = \langle \Box\neg(p\wedge\neg p): -- \rangle$
\end{example1}
\noindent Arguments $a_5$ and $a_6$ are incoherent in the sense that in constructing them we relied on arguments the conclusions of which are conflicting (namely $a_1$ and $a_2$). It is vital that we are able to filter out such incoherent arguments. Definition \ref{def_fattack} takes care of that. By attacking $a_6$, argument $a_7$ protects (defends) the unproblematic $a_3$, which is attacked by $a_6$ in view of Definition \ref{def_cattack} below. We return to this point in footnote \ref{fn_inc}, after we explained how arguments are evaluated.

The second type of attack relation ensures that mutually incompatible obligations attack each other:
\begin{definition1}[Conflict attack]\label{def_cattack}
$a=\langle \OOs {-} A : \ldots \rangle$ attacks $b=\langle \OOs A: \ldots \rangle$, and $a$ attacks all of $b$'s super-arguments.
\end{definition1}
In Example \ref{ex_ctd}, arguments $a_3$ and $a_4$ attack each other according to Definition \ref{def_cattack}. Moreover, $a_3$ attacks $a_5$ and $a_6$; and $a_4$ attacks $a_5$. Likewise, in Example \ref{ex_incoh}, $a_1$ and $a_2$ attack each other, and so do $a_3$ and $a_6$. Moreover, $a_1$ attacks $a_5$ and $a_6$; and $a_2$ attacks $a_4, a_5$, and $a_6$.

\begin{example1}[Conflict attack]\label{ex_spec2}
Let $\Gamma_{\Gtel\label{gspec2}} = \{p,q, p\Ra r, (p\wedge q)\Ra s, \Box\neg(r\wedge s)\}$. We construct the following arguments on the basis of $\Gamma_{\ref{gspec2}}$:\smallskip

\noindent\begin{tabular}{l l l l}
$a_1$: & $\langle \OOs r: p, p\Ra r\rangle$ & \quad$a_4$: & $\langle\Box\neg(r\wedge s): -- \rangle$ \\
$a_2$: & $\langle \OOs s: p\wedge q, (p\wedge q)\Ra s\rangle$ & \quad$a_5$: & $\langle \OOs\neg r: a_2, \Box(s\supset\neg r) \rangle$ \\
$a_3$: & $\langle \OOs(r\wedge s): a_1,a_2 \rangle$ & \quad$a_6$: & $\langle \OOs\neg s: a_1, \Box(r\supset\neg s)\rangle$\\
\end{tabular}\smallskip

\noindent $a_4$ attacks $a_3$ by Definition \ref{def_fattack}. By Definition \ref{def_cattack} $a_1$ attacks $a_5$; $a_5$ attacks $a_1, a_3$, and $a_6$; $a_2$ attacks $a_6$; and $a_6$ attacks $a_2, a_3$, and $a_5$.
\end{example1}


\subsection{Evaluating deontic arguments}\label{sub_deval}

For the evaluation of deontic arguments relative to a premise set, we extend Dung-style AFs to deontic argumentation frameworks, and we borrow Dung's argument evaluation mechanism from Definitions \ref{def_af}-\ref{def_ground2}:
\begin{definition1}[DAF]\label{def_daf}
The \emph{deontic argumentation framework} (DAF) for $\Gamma\subseteq\mathcal{L}^P\cup\mathcal{L}^{\Box}\cup\mathcal{L}^{\Ra}$ is an ordered pair $\langle \mathcal{A}(\Gamma), \mathsf{Att}(\Gamma) \rangle$ where
\begin{itemize}[itemsep=-1mm]
\item $\mathcal{A}(\Gamma)$ is the set of arguments constructed from $\Gamma$ in line with Definition \ref{def_arg}; and
\item where $a,b\in\mathcal{A}(\Gamma
)$: $(a,b)\in\mathsf{Att}(\Gamma)$ iff $a$ attacks $b$ according to Definition \ref{def_fattack} or Definition \ref{def_cattack}.
\end{itemize}
\end{definition1}
Like AFs, DAFs can be represented as directed graphs. Here, for instance, is a graph depicting the arguments we constructed on the basis of $\Gamma_{\ref{gctd}}$:\footnote{Due to space limitations, we leave it to the reader to construct similar graphs for the other examples in this paper.}

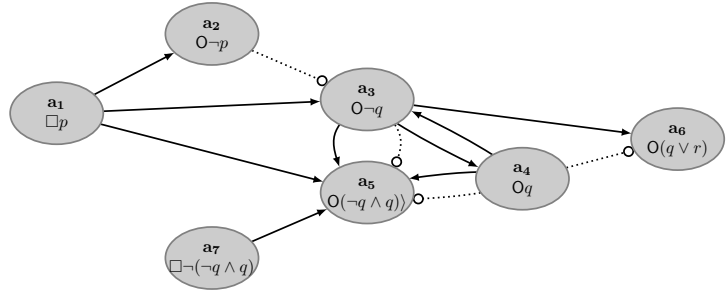
\begin{figure}[h]
\centering
\scalebox{.65}{
\begin{tikzpicture}[>=latex,line join=bevel,]
  \pgfsetlinewidth{1bp}
\begin{scope}
  \pgfsetstrokecolor{black}
  \definecolor{strokecol}{rgb}{1.0,1.0,1.0};
  \pgfsetstrokecolor{strokecol}
  \definecolor{fillcol}{rgb}{1.0,1.0,1.0};
  \pgfsetfillcolor{fillcol}
  \filldraw (0.0bp,0.0bp) -- (0.0bp,166.0bp) -- (414.0bp,166.0bp) -- (414.0bp,0.0bp) -- cycle;
\end{scope}
  \pgfsetcolor{black}
  \draw [->,solid] (54.265bp,103.18bp) .. controls (84.851bp,104.56bp) and (135.55bp,106.83bp)  .. (179.92bp,108.83bp);
  \draw [->,dotted,-o] (140.21bp,138.41bp) .. controls (150.63bp,133.91bp) and (163.23bp,128.47bp)  .. (183.8bp,119.59bp);
  \draw [->,solid] (279.67bp,77.92bp) .. controls (269.02bp,84.717bp) and (254.72bp,92.59bp)  .. (232.51bp,103.31bp);
  \draw [->,dotted,-o] (322.51bp,70.407bp) .. controls (331.57bp,72.774bp) and (342.01bp,75.503bp)  .. (361.62bp,80.628bp);
  \draw [->,dotted,-o] (272.71bp,55.847bp) .. controls (263.83bp,54.38bp) and (253.5bp,53.238bp)  .. (233.45bp,52.128bp);
  \draw [->,solid] (270.56bp,67.872bp) .. controls (261.42bp,67.632bp) and (250.97bp,66.809bp)  .. (231.12bp,64.125bp);
  \draw [->,solid] (52.377bp,95.7bp) .. controls (83.314bp,87.705bp) and (136.86bp,73.869bp)  .. (181.63bp,62.297bp);
  \draw [->,solid] (224.55bp,95.936bp) .. controls (235.21bp,89.155bp) and (249.44bp,81.325bp)  .. (271.54bp,70.666bp);
  \draw [->,dotted,-o] (223.06bp,95.287bp) .. controls (226.46bp,90.306bp) and (227.79bp,85.325bp)  .. (223.09bp,70.766bp);
  \draw [->,solid] (190.94bp,95.287bp) .. controls (187.54bp,90.306bp) and (186.21bp,85.325bp)  .. (190.91bp,70.766bp);
  \draw [->,solid] (233.88bp,106.66bp) .. controls (264.55bp,102.69bp) and (315.83bp,96.067bp)  .. (360.37bp,90.311bp);
  \draw [->,solid] (140.21bp,27.589bp) .. controls (150.63bp,32.089bp) and (163.23bp,37.532bp)  .. (183.8bp,46.414bp);
  \draw [->,solid] (48.855bp,112.9bp) .. controls (60.033bp,118.74bp) and (73.998bp,126.04bp)  .. (95.295bp,137.18bp);
\begin{scope}
  \definecolor{strokecol}{rgb}{0.0,0.0,0.0};
  \pgfsetstrokecolor{strokecol}
  \draw [hlund] (117.0bp,18.0bp) ellipse (27.0bp and 18.0bp);
  \draw (117.0bp,18.0bp) node {$\array{c}{\bf a_7}\\ \Box\neg(\neg q\wedge q) \endarray$};
\end{scope}
\begin{scope}
  \definecolor{strokecol}{rgb}{0.0,0.0,0.0};
  \pgfsetstrokecolor{strokecol}
  \draw [hlund] (387.0bp,87.0bp) ellipse (27.0bp and 18.0bp);
  \draw (387.0bp,87.0bp) node {$\array{c}{\bf a_6}\\ {\sf O} (q\vee r) \endarray$};
\end{scope}
\begin{scope}
  \definecolor{strokecol}{rgb}{0.0,0.0,0.0};
  \pgfsetstrokecolor{strokecol}
  \draw [hlund] (207.0bp,56.0bp) ellipse (27.0bp and 18.0bp);
  \draw (207.0bp,56.0bp) node {$\array{c}{\bf a_5}\\ {\sf O} (\neg q\wedge q) \rangle\endarray$};
\end{scope}
\begin{scope}
  \definecolor{strokecol}{rgb}{0.0,0.0,0.0};
  \pgfsetstrokecolor{strokecol}
  \draw [hlund] (297.0bp,64.0bp) ellipse (27.0bp and 18.0bp);
  \draw (297.0bp,64.0bp) node {$\array{c}{\bf a_4}\\ {\sf O} q \endarray$};
\end{scope}
\begin{scope}
  \definecolor{strokecol}{rgb}{0.0,0.0,0.0};
  \pgfsetstrokecolor{strokecol}
  \draw [hlund] (207.0bp,110.0bp) ellipse (27.0bp and 18.0bp);
  \draw (207.0bp,110.0bp) node {$\array{c}{\bf a_3}\\ {\sf O} \neg q \endarray$};
\end{scope}
\begin{scope}
  \definecolor{strokecol}{rgb}{0.0,0.0,0.0};
  \pgfsetstrokecolor{strokecol}
  \draw [hlund] (117.0bp,148.0bp) ellipse (27.0bp and 18.0bp);
  \draw (117.0bp,148.0bp) node {$\array{c}{\bf a_2}\\ {\sf O} \neg p \endarray$};
\end{scope}
\begin{scope}
  \definecolor{strokecol}{rgb}{0.0,0.0,0.0};
  \pgfsetstrokecolor{strokecol}
  \draw [hlund] (27.0bp,102.0bp) ellipse (27.0bp and 18.0bp);
  \draw (27.0bp,102.0bp) node {$\array{c}{\bf a_1}\\ \Box p\endarray$};
\end{scope}
\end{tikzpicture}
}
\caption{Arguments and attack relations for $\Gamma_{\ref{gctd}}$.}
\end{figure}

Nodes in the graph represent the arguments constructed on the basis of $\Gamma_{\ref{gctd}}$ in Example \ref{ex_ctd}. Below the arguments' names, we stated their conclusions. Arrows represent attacks. Dotted lines represent sub-argument relations.

We evaluate arguments in a DAF using Dung's grounded semantics from Section \ref{sub_abstr}: In Definition \ref{def_af}, replace $\mathcal{A}$ (resp.\ $\mathsf{Att}$) with $\mathcal{A}(\Gamma)$ (resp.\ $\mathsf{Att}(\Gamma)$). Similarly for Definition \ref{def_ground2}, where we also replace occurrences of $\mathcal{G}$ and $\mathcal{G}_i$ with $\mathcal{G}(\Gamma)$ and $\mathcal{G}_i(\Gamma)$ respectively.

Let us now apply Definition \ref{def_ground2} to Example \ref{ex_ctd}. Clearly, $a_1,a_7 \in \mathcal{G}_0 (\Gamma_{\ref{gctd}})$, since Definitions \ref{def_fattack} and \ref{def_cattack} provide us with no means to attack arguments the conclusions of which are members of $\Gamma_{\ref{gctd}}^{\Box}$. In the next step of our construction, $a_4, a_6 \in\mathcal{G}_1(\Gamma_{\ref{gctd}})$, since they are defended by $a_1\in\mathcal{G}_0 (\Gamma_{\ref{gctd}})$. $a_2, a_3, a_5\not\in\mathcal{G}_1(\Gamma_{\ref{gctd}})$, since each of these arguments is attacked by $a_1$ (hence undefended).

We cannot construct any further arguments which attack $a_4$ or $a_6$ and which do not contain any of the undefended arguments $a_2$ or $a_3$ as sub-arguments. Moreover, we show in the Appendix (Lemma \ref{lem:G:G1}) that, for any premise set $\Gamma$, if $a\in\mathcal{G}(\Gamma)$, then $a\in\mathcal{G}_1(\Gamma)$. By the Definition \ref{def_ground2}, $a_1,a_4,a_6,a_7\in\mathcal{G}(\Gamma_{\ref{gctd}})$ while $a_2,a_3,a_5\not\in\mathcal{G}(\Gamma_{\ref{gctd}})$.


\begin{definition1}[DAF-consequence]\label{def_dafcons}
Where $\Gamma\subseteq\mathcal{L}^P\cup\mathcal{L}^{\Box}\cup\mathcal{L}^{\Ra}$ and $A\in\mathcal{L}^P$, $\Gamma\vdash_{\sys{DAF}}\OOs A$ iff there is an argument $a\in\mathcal{G}(\Gamma)$ with conclusion $\OOs A$. 
\end{definition1}
\noindent By Definition \ref{def_dafcons}, $\Gamma_{\ref{gctd}}\vdash_{\sys{DAF}} \OOs q$ and $\Gamma_{\ref{gctd}}\vdash_{\sys{DAF}} \OOs (q\vee r)$, while $\Gamma_{\ref{gctd}} \not\vdash_{\sys{DAF}} \OOs \neg p$ and $\Gamma_{\ref{gctd}}\not\vdash_{\sys{DAF}} \OOs \neg q$.

\noindent In Example \ref{ex_incoh}, $\Gamma_{\ref{ginc}} \vdash_{\sys{DAF}} \OOs q$.\footnote{\label{fn_inc} The conclusion $\OOs q$ of argument $a_3$ in Example \ref{ex_incoh} is accepted despite its being attacked by $a_6$. The reason is that $a_6$ is in turn attacked by $a_7$, so that $a_7$ defends $a_3$ from the attack by $a_6$.}
We leave it to the reader to check that none of $\OOs p, \OOs\neg p, \OOs(p\vee\neg q)$, or $\OOs\neg q$ is a \sys{DAF}-consequence of $\Gamma_{\ref{ginc}}$, and that none of $\OOs r, \OOs s, \OOs (r\wedge s), \OOs\neg r$, or $\OOs \neg s$ is a \sys{DAF}-consequence of $\Gamma_{\ref{gspec2}}$.


\subsection{Rationality postulates}\label{sub_meta}

In \cite[Sec.~4]{CamAmg07} the properties of output closure and output consistency were proposed as desiderata for well-behaved argumentation systems. Where $Output(\Gamma) = \{A\mid \Gamma\vdash_{\sys{DAF}}\OOs A\}$:
\begin{property}[Closure]\label{prop_closure}
$Output(\Gamma) = \Cn{CL}{}{Output(\Gamma)}$.
\end{property}
\begin{property}[Consistency]\label{prop_cons}
$\Cn{CL}{}{Output(\Gamma)}$ is consistent.
\end{property}
Properties \ref{prop_closure} and \ref{prop_cons} follow for \sys{DAF} in view of resp.\ Theorems \ref{thm:closure} and \ref{thm:consistency} in the Appendix. Property \ref{prop_cautcut} is proven in Theorem \ref{thm:CT} in the Appendix:
\begin{property}[Cautious cut/cumulative transitivity]\label{prop_cautcut}
Let $\Delta_{\Ra} = \{\top\Ra A\mid A\in\Delta\}$. If $\Gamma\vdash_{\sys{DAF}}\OOs A$ for all $A\in\Delta$ and $\Gamma\cup\Delta_{\Ra}\vdash_{\sys{DAF}}\OOs B$, then $\Gamma\vdash_{\sys{DAF}}\OOs B$.
\end{property}
Properties \ref{prop_cautmon} and \ref{prop_ratmon} fail for \sys{DAF}:
\begin{property}[Cautious monotonicity]\label{prop_cautmon}
If $\Gamma\vdash_{\sys{DAF}}\OOs A$ and $\Gamma\vdash_{\sys{DAF}} \OOs B$, then $\Gamma\cup\{\top\Ra A\}\vdash_{\sys{DAF}}\OOs B$.
\end{property}
\begin{property}[Rational monotonicity]\label{prop_ratmon}
If $\Gamma\vdash_{\sys{DAF}}\OOs A$ and $\Gamma\not\vdash_{\sys{DAF}}\OOs\neg B$, then $\Gamma\cup\{\top\Ra B\}\vdash_{\sys{DAF}}\OOs A$
\end{property}

\begin{example1}[Failure of properties \ref{prop_cautmon} and \ref{prop_ratmon}, adapted from \cite{Cam04hy}]\label{ex_failmeta}
Let $\Gamma_{\Gtel\label{gcautmon}} = \{p, p\Ra q, q\Ra r, r\Ra\neg q,\neg q\Ra s,\top\Ra\neg s\}$. We construct the following arguments on the basis of $\Gamma_{\ref{gcautmon}}$:\smallskip

\begin{tabular}{l l l l}
$a_1$: & $\langle \OOs q: p, p\Ra q\rangle$ & $a_4$: & $\langle \OOs s: a_3, \neg q\Ra s\rangle$ \\
$a_2$: & $\langle \OOs r: a_1, q\Ra r \rangle$ & $a_5$: & $\langle \OOs\neg s: \top, \top\Ra\neg s\rangle$ \\
$a_3$: & $\langle \OOs\neg q: a_2, r\Ra\neg q \rangle$ & $a_6$: & $\langle \Box\neg(q\wedge\neg q): --\rangle$ \\
\end{tabular}\smallskip

By Definition \ref{def_cattack}: $a_1$ attacks $a_3$ and $a_4$; $a_3$ attacks all of $a_1$-$a_4$ (including itself); and $a_4$ and $a_5$ attack each other. By Definition \ref{def_fattack}, $a_6$ attacks $a_3$ and $a_4$, since both $q$ and $\neg q$ are members of $\UUs\OOs(a_3)$ and $\UUs\OOs(a_4)$. As a result, $\OOs q,\OOs r$, and $\OOs \neg s$ are \sys{DAF}-consequences of $\Gamma_{\ref{gcautmon}}$, while $\OOs\neg q$ and $\OOs s$ are not.

Now add the new conditional obligation $\top\Ra r$ to $\Gamma_{\ref{gcautmon}}$, so that we obtain the new arguments\smallskip

\begin{tabular}{l l l l}
$a_7$: & $\langle \OOs r: \top, \top\Ra r\rangle$ & $a_9$: & $\langle \OOs s: a_8, \neg q\Ra s\rangle$ \\
$a_8$: & $\langle \OOs\neg q: a_7, r\Ra\neg q \rangle$ & \ & \ \\
\end{tabular}\smallskip

None of these new arguments is attacked by $a_6$, which defends $a_1$ and $a_5$ from the attacks by $a_3$ and $a_4$ respectively. By Definition \ref{def_cattack}, $a_8$ and $a_1$ attack each other. So do $a_9$ and $a_5$. As a result, none of $a_1, a_5, a_8$, and $a_9$ is in the grounded extension of $\Gamma_{\ref{gcautmon}}\cup\{\top\Ra r\}$. So we have a counter-example to Property \ref{prop_cautmon}: $\Gamma_{\ref{gcautmon}}\vdash_{\sys{DAF}}\OOs r$ and $\Gamma_{\ref{gcautmon}}\vdash_{\sys{DAF}}\OOs\neg s$, while $\Gamma_{\ref{gcautmon}}\cup\{\top\Ra r\}\not\vdash_{\sys{DAF}}\OOs\neg s$.
\end{example1}
This example also serves to illustrate the failure of Property \ref{prop_ratmon} for \sys{DAF}. Arguments with conclusion $\OOs\neg r$ can be constructed on the basis of $\Gamma_{\ref{gcautmon}}$ only on the basis of incoherent arguments. Let, for instance:\smallskip

\begin{tabular}{l l l l}
$a_{10}$: & $\langle \OOs (q\wedge\neg q): a_1,a_3\rangle$ & $a_{11}$: & $\langle \OOs \neg r: a_{10}, \Box((q\wedge\neg q)\supset\neg r)\rangle$ \\
\end{tabular}\smallskip

In view of Definition \ref{def_fattack}, arguments constructed on an incoherent basis are attacked by an otherwise unattacked argument. For instance, $a_{11}$ is attacked by the unattacked argument $a_6$. Because of this, $\Gamma_{\ref{gcautmon}}\not\vdash_{\sys{DAF}}\OOs\neg r$. But then, since $\Gamma_{\ref{gcautmon}}\vdash_{\sys{DAF}}\OOs\neg s$ and $\Gamma_{\ref{gcautmon}}\cup\{\top\Ra r\}\not\vdash_{\sys{DAF}}\OOs\neg s$, Property \ref{prop_ratmon} fails for \sys{DAF}.


\section{Beyond the basics}\label{sec_extensions}

\subsection{Conflict-resolution}\label{sub_confl}

\subsubsection{Resolving conflicts via logical analysis}\label{sub_spec}

It has been argued that, in cases of conflict, more specific obligations should be given precedence over less specific ones.\footnote{Understood in this way, specificity cases have been studied extensively in the fields of non-monotonic logic (see e.g.\ \cite{DunSon01,DelSch97}) and deontic logic (see e.g.\ \cite{CarJon02,PraSer96,DBLP:journals/japll/Strasser11,strasser2015normative}).} Consider the following example:
\begin{example1}[Specificity]\label{ex_celery}
Let $\Gamma_{\Gtel\label{gcelery}} = \{q,r, q\Ra p, (q\wedge r)\Ra\neg p\}$. We can interpret $\Gamma_{\ref{gcelery}}$ as representing a scenario in which an agent is making carrot soup. Let $p,q$, and, respectively, $r$ abbreviate `there is fennel', `there are carrots', and `there is celery'. If there are carrots in the garden still, our agent should take care that he buys fennel in order to make the soup ($q\Ra p$). However, if both carrots and celery are in the garden, he should not get fennel ($(q\wedge r)\Ra\neg p$), because celery can be used instead of fennel. As it turns out, both carrots and celery are in his garden ($q,r$). The desirable outcome in this case is that the agent ought not go out and buy fennel.
\end{example1}
A principled way of obtaining outcomes in which more specific obligations are preferred over less specific ones, is to define specificity in terms of logical strength, and to define a new attack relation for letting more specific arguments attack less specific ones. Let the \emph{factual support} of a deontic argument $a$ be the set $\SSs(a) = \{B\mid B\in(\Cs(a)\cap \mathcal{L}^P)\}$.

We write $\SSs(a)\sqsubseteq \SSs(b)$ iff for all $A\in\SSs(a)$ there is a $B\in\SSs(b)$ such that $A\vdash B$ and for all $B\in\SSs(b)$ there is an $A\in\SSs(a)$ such that $A\vdash B$.
$\SSs(a)\sqsubset \SSs(b)$ ($a$ is \emph{more specific} than $b$) iff $\SSs(a)\sqsubseteq \SSs(b)$ and $\SSs(b)\not\sqsubseteq \SSs(a)$.

We replace Definition \ref{def_cattack} with Definition \ref{def_specattack}:
\begin{definition1}[Conflict attack w/specificity]\label{def_specattack} Let $a = \langle \OOs {-} A : \ldots \rangle$  and $b = \langle \OOs A: \ldots \rangle$.
\begin{itemize}
\item[(i)] If $\SSs(a)\sqsubset \SSs(b)$, then $a$ attacks $b$ and all of $b$'s super-arguments,
\item[(ii)] $b$ attacks $a$ and all of $a$'s super-arguments, unless $a$ attacks $b$ in view of clause (i).
\end{itemize}
\end{definition1}
Let \sys{DAF_s} (with subscript `s' for specificity) be the logic resulting from constructing the attack relation $\mathsf{Att}$ on the basis of Definitions \ref{def_fattack} and \ref{def_specattack}.

In Example \ref{ex_celery}, we construct the following arguments from $\Gamma_{\ref{gcelery}}$:\smallskip

\noindent\begin{tabular}{l l}
$a_1$: & $\langle \OOs p: q, q\Ra p\rangle$\\
$a_2$: & $\langle \OOs\neg p: q\wedge r, (q\wedge r)\Ra\neg p\rangle$\\
\end{tabular}\smallskip

Since $\SSs(a_2)\sqsubset\SSs(a_1)$, $a_2$ attacks $a_1$ by Definition \ref{def_specattack}, but not vice versa. As a result, only $a_2$ is in $\Gamma_{\ref{gcelery}}$'s grounded extension, and $\Gamma_{\ref{gcelery}}\vdash_{\sys{DAF_s}}\OOs \neg p$, while $\Gamma_{\ref{gcelery}}\not\vdash_{\sys{DAF_s}}\OOs p$.

In Example \ref{ex_spec2}, the factual support of the arguments constructed from $\Gamma_{\ref{gspec2}}$ is such that $\SSs(a_2)=\SSs(a_5) \sqsubset \SSs(a_1) = \SSs(a_6)$. By Definition \ref{def_specattack}, $a_5$ attacks $a_1$ and $a_2$ attacks $a_6$. As a result, the more specific arguments $a_2$ and $a_5$ defeat the less specific $a_1$ and $a_6$, so that $\Gamma_{\ref{gspec2}}\vdash_{\sys{DAF_s}}\OOs s$ and $\Gamma_{\ref{gspec2}}\vdash_{\sys{DAF_s}}\OOs \neg r$, while $\Gamma_{\ref{gspec2}}\not\vdash_{\sys{DAF_s}}\OOs r$ and $\Gamma_{\ref{gspec2}}\not\vdash_{\sys{DAF_s}}\OOs \neg s$. As before, $\Gamma_{\ref{gspec2}}\not\vdash_{\sys{DAF_s}}\OOs (r\wedge s)$.

In dealing with conflict-resolution via logical analysis, we have chosen for a cautious notion of specificity. For instance, $\{p\}\not\sqsubset\{p,q\}$ and $\{p\}\not\sqsubset\{p\wedge q,r\}$. In certain contexts it may be sensible to opt for a stronger characterization of `$\sqsubset$'. A detailed discussion of such issues would lead us too far astray given our present purposes. Instead, we point out that our framework readily accommodates alternative characterizations of `$\sqsubset$' to be used in Definition \ref{def_specattack}.

\subsubsection{Resolving conflicts via priorities}

Instead of (or in combination with) conflict-resolution via logical analysis, a priority ordering $\leq$ can be introduced over conditional norms, and our formal language can be adjusted accordingly. Conditional norms then come with an associated degree of priority $\alpha\in\mathbb{Z}^+$, written $A\Ra_{\alpha} B$ (higher numbers denote higher priorities).

We lift $\leq$ to a priority ordering $\preceq$ over arguments via the \emph{weakest link} principle: an argument is only as strong as the weakest priority conditional used in its construction \cite{Pra11}. Let $\mathsf{Pr}(\Delta)=\{\alpha\mid A\Ra_{\alpha}B\in\Delta\}$ and let $min(\mathsf{Pr}(\Delta))$ be the lowest $\alpha\in\mathsf{Pr}(\Delta)$. Then $\Delta\preceq\Delta'$ iff $min(\mathsf{Pr}(\Delta)) \leq min(\mathsf{Pr}(\Delta'))$. Relative to a premise set $\Gamma$, we write $a\preceq b$ iff $\Cs(a)\cap\Gamma^{\Ra} \preceq \Cs(b)\cap\Gamma^{\Ra}$. $a\prec b$ iff $a\preceq b$ and $b\not\preceq a$.

We replace Definition \ref{def_cattack} with the following definition:
\begin{definition1}[Prioritized conflict attack]\label{def_pattack}
If $a \not\prec b$, then $a=\langle\OOs {-}A:\ldots\rangle$ attacks $b=\langle\OOs A:\ldots\rangle$ and all of $b$'s super-arguments.
\end{definition1}
Let \sys{DAF_{\leq}} be the logic resulting from constructing the attack relation $\mathsf{Att}$ on the basis of Definitions \ref{def_fattack} and \ref{def_pattack}.
\begin{example1}[Prioritized conflict attack]\label{ex_3spec}
Let $\Gamma_{\Gtel\label{gspec3}} = \{p,q,r,\Box\neg(s\wedge t\wedge u), p\Ra_1 s, q\Ra_2 t, r\Ra_3 u\}$. We construct the following arguments on the basis of $\Gamma_{\ref{gspec3}}$:\smallskip

\noindent\begin{tabular}{l l l l}
$a_1$: & $\langle \Box\neg(s\wedge t\wedge u): -- \rangle$ & $a_8$: & $\langle \OOs(s\wedge t\wedge u): a_4, a_5 \rangle$ \\
$a_2$: & $\langle \OOs s: p, p\Ra_1 s \rangle$ & $a_9$: & $\langle \OOs\neg(t\wedge u): a_2, \Box(s\supset\neg(t\wedge u)) \rangle$ \\
$a_3$: & $\langle \OOs t: q, q\Ra_2 t\rangle$ & $a_{10}$: & $\langle \OOs\neg(s\wedge u): a_3, \Box(t\supset\neg(s\wedge u)) \rangle$ \\
$a_4$: & $\langle \OOs u: r, r\Ra_3 u \rangle$ & $a_{11}$: & $\langle \OOs\neg(s\wedge t): a_4, \Box(u\supset\neg(s\wedge t)) \rangle$ \\
$a_5$: & $\langle \OOs (s\wedge t): a_2, a_3\rangle$ & $a_{12}$: & $\langle \OOs\neg u: a_5, \Box((s\wedge t)\supset\neg u) \rangle$ \\
$a_6$: & $\langle \OOs (s\wedge u): a_2, a_4\rangle$ & $a_{13}$: & $\langle \OOs\neg t: a_6, \Box((s\wedge u)\supset\neg t) \rangle$ \\
$a_7$: & $\langle \OOs (t\wedge u): a_3, a_4\rangle$ & $a_{14}$: & $\langle \OOs\neg s: a_7, \Box((t\wedge u)\supset\neg s) \rangle$ \\
\end{tabular}\smallskip

The order of arguments is such that $a_{2},a_{5},a_{6},a_{8},a_{9},a_{12},a_{13} \prec a_{3},a_{7},a_{10},a_{14} \prec a_{4},a_{11}$. By Definition \ref{def_pattack}, $a_{14}$ attacks $a_{2}, a_{5}$, $a_{6}$, $a_{8}$, $a_{9}$, $a_{12}$, and $a_{13}$; $a_{3}$ attacks $a_{13}$; $a_{4}$ attacks $a_{12}$; $a_{11}$ attacks $a_{5}, a_8$, and $a_{12}$; $a_{10}$ attacks $a_{6}$ and $a_{13}$; and $a_{7}$ attacks $a_{9}$. By Definition \ref{def_fattack}, $a_{1}$ attacks $a_{8}$.
As a result, $a_{1},a_{3},a_{4},a_{7},a_{10},a_{11},a_{14}\in\mathcal{G}(\Gamma_{\ref{gspec3}})$, while $a_{2},a_{5},a_{6},a_{8},a_{9},a_{12},a_{13}\not\in\mathcal{G}(\Gamma_{\ref{gspec3}})$. The following obligations are \sys{DAF_{\leq}}-consequences of $\Gamma_{\ref{gspec3}}: \OOs t, \OOs u, \OOs (t\wedge u), \OOs\neg(s\wedge u), \OOs\neg(s\wedge t), \OOs\neg s$. The following obligations are not \sys{DAF_{\leq}}-derivable from $\Gamma_{\ref{gspec3}}: \OOs s, \OOs(s\wedge t), \OOs(s\wedge u), \OOs(s\wedge t\wedge u), \OOs\neg(t\wedge u), \OOs\neg u, \OOs\neg t$.
\end{example1}
As with `$\sqsubset$' in Definition \ref{def_specattack}, there are other ways of characterizing `$\prec$' in Definition \ref{def_pattack}. For instance, instead of lifting $\leq$ via the weakest link principle, we could lift it via the \emph{strongest link} principle, according to which an argument is as strong as the strongest priority conditional used in its construction.\footnote{If the strongest link principle is used, Definition \ref{def_pattack} should no longer allow for attacks on super-arguments, since $a\not\prec b$ no longer warrants that $a\not\prec c$ where $c$ is a super-argument of $b$.
A further alternative is to use the \emph{last link} principle, according to which an argument gets the priority of the conditional which occurs last in its proof sequence. 
} Depending on the way $\leq$ is lifted to $\preceq$, different outcomes are possible with respect to the priority puzzles studied in e.g.\ \cite{Han08,Hor07,Hor12}. A thorough investigation of these puzzles within our framework is left for an extended version of this paper.

\subsection{Anticipating violations and conflicts}\label{sub_shadow}
Obligations which are violated or conflicted should not be detached. But what about obligations that \emph{commit} us to violations or conflicts? Consider the following example, adapted from \cite{Mak94,MakTor01}.
\begin{example1}\label{ex_abc}
Let $\Gamma_{\Gtel\label{gbday}} = \{p, p\Ra q, q\Ra r, r\Ra\neg p\}$. We construct the following arguments on the basis of $\Gamma_{\ref{gbday}}$:\smallskip

\noindent\begin{tabular}{l l l l}
$a_1$: & $\langle \Box p: -- \rangle$ & \quad\quad$a_3$: & $\langle \OOs r: a_2, q\Ra r\rangle$\\
$a_2$: & $\langle \OOs q: p, p\Ra q\rangle$ & \quad\quad$a_4$: & $\langle \OOs\neg p: a_3, r\Ra\neg p\rangle$\\
\end{tabular}\smallskip

Suppose you are throwing a party. Let $p$ (resp.\ $q,r$) abbreviate `Peggy (resp.\ Quincy, Ruth) is invited to the party'. If Peggy is invited, then Quincy should be invited as well (perhaps because they are good friends and we know both of them). Likewise, if Quincy is invited then Ruth should be invited as well. But if Ruth is invited, then Peggy should not be (perhaps because we know Ruth and Peggy do not get along well). It is settled that Peggy is invited. You already sent her the official invitation, and it would be too awkward to tell her she can't come. Should Quincy and/or Ruth be invited?
\end{example1}
Arguments $a_1,a_2$, and $a_3$ are in $\Gamma_{\ref{gbday}}$'s grounded extension $\mathcal{G}(\Gamma_{\ref{gbday}})$. $a_4$ is not in $\mathcal{G}(\Gamma_{\ref{gbday}})$ since it is attacked by $a_1$ according to Definition \ref{def_fattack}; consequently, $\Gamma_{\ref{gbday}}\vdash_{\sys{DAF}}\OOs q$ and $\Gamma_{\ref{gbday}}\vdash_{\sys{DAF}}\OOs r$, while $\Gamma_{\ref{gbday}}\not\vdash_{\sys{DAF}}\OOs\neg p$.

A more cautious reasoner may argue that $\OOs q$ and $\OOs r$ should not be detached, since they lead to a commitment to $\OOs\neg p$: they form part of the detachment chain of $a_4$. This commitment reflects very badly on arguments $a_2$ and $a_3$, since $\OOs \neg p$ is violated.

To model this behavior, we introduce the \emph{deontic doubt operator} $\odot$. We will use this operator to construct new arguments, called \emph{shadow arguments}, the conclusion of which is of the form $\odot A$. A shadow argument with conclusion $\odot A$ casts doubt on -- and attacks -- arguments with conclusion $\OOs A$. Shadow arguments cannot be used to support obligations, but only to attack other arguments. They can \emph{only} rule out deontic arguments. They cannot generate new consequences.\footnote{Shadow arguments are similar in spirit to Caminada's \emph{HY-arguments} from \cite{Cam04hy}. An HY-argument $a$ is an incoherent argument constructed on the basis of the conclusion of another argument $b$. Since $a$ shows that $b$ leads to incoherence, $b$'s conclusion is attacked by the HY-argument $a$. Caminada shows how in the presence of HY-arguments, the property of cautious monotonicity may be restored for AFs. The same holds true for shadow arguments in our setting (cfr.\ infra). As Caminada's construction is defined within a framework consisting only of literals and (defeasible) rules relating (conjunctions of) literals, we cannot employ it in our setting.}

In the resulting system \sys{DAF_{\odot}}, our language $\mathcal{L}$ is adjusted so as to include members of $\mathcal{P}$ within the scope of the new operator $\odot$. Arguments are constructed in line with Definition \ref{def_doubtarg}:
\begin{definition1}\label{def_doubtarg}
Given a premise set $\Gamma$, we allow rules (i)-(vii) for constructing arguments, where (i)-(v) are the rules from Definition \ref{def_arg}:
\begin{itemize}[itemsep=-1mm]
\item[(vi)] If $a=\langle \Box A: -- \rangle$ is an argument, then $\langle \odot{-}A: a\rangle$ is an argument;
\item[(vii)] If $a=\langle \OOs A: \ldots \rangle$ is an argument, then $\langle \odot{-}A: a\rangle$ is an argument.
\end{itemize}
\end{definition1}
We say that an argument $a$ has \emph{minimal support} if there is no argument $b$ with the same conclusion such that $\mathsf{C}(b) \subset \mathsf{C}(a)$. In \sys{DAF_{\odot}} the attack relation is constructed on the basis of Definition \ref{def_shattack}:\footnote{By the construction of Definition \ref{def_shattack}, Definitions \ref{def_fattack} and \ref{def_cattack} become redundant in \sys{DAF_{\odot}}. All cases covered by these definitions are covered already by Definition \ref{def_shattack}.}
\begin{definition1}[Shadow attack]\label{def_shattack} Where $a = \langle \OOs A: \ldots \rangle$ has minimal support:
\begin{itemize}
\item[(i)] Where $b$ is a deontic sub-argument of $a$, $\langle \odot A: \ldots\rangle$ attacks $b$ as well as all of $b$'s super-arguments,
\item[(ii)] Where $b$ is a deontic sub-argument of $a$ and $\emptyset\neq\Theta\subseteq \UUs\OOs(a)$, $\langle \odot \bigwedge\Theta: \ldots\rangle$ attacks $b$ as well as all of $b$'s super-arguments.
\end{itemize}
\end{definition1}
Reconsider $\Gamma_{\ref{gbday}}$ from Example \ref{ex_abc}. From $a_1$, we can construct the shadow argument $a_5 = \langle \odot\neg p: a_1\rangle$. By clause (i) of Definition \ref{def_shattack}, $a_5$ attacks $a_4, a_3$, and $a_2$. As a result, $a_2$ and $a_3$ are no longer in $\mathcal{G}(\Gamma_{\ref{gbday}})$. $\Gamma_{\ref{gbday}}\not\vdash_{\sys{DAF_{\odot}}}\OOs q$ and  $\Gamma_{\ref{gbday}}\not\vdash_{\sys{DAF_{\odot}}}\OOs r$.
\begin{example1}\label{ex_shadow}
Let $\Gamma_{\Gtel\label{gsha}} = \{\Box s, \top\Ra p, \top\Ra q, (p\wedge q)\Ra r, r\Ra\neg s, q\Ra t\}$. We construct the following arguments on the basis of $\Gamma_{\ref{gsha}}$:\smallskip

\noindent\begin{tabular}{l l l l}
$a_1$: & $\langle\Box s: -- \rangle$ & \quad\quad$a_5$: & $\langle\OOs r: a_4, (p\wedge q)\Ra r \rangle$\\
$a_2$: & $\langle\OOs p: \top, \top\Ra p \rangle$ & \quad\quad$a_6$: & $\langle\OOs\neg s: a_5, r\Ra\neg s \rangle$\\
$a_3$: & $\langle\OOs q: \top, \top\Ra q \rangle$ & \quad\quad$a_7$: & $\langle \OOs t: a_3, q\Ra t \rangle$\\
$a_4$: & $\langle\OOs(p\wedge q): a_2,a_3 \rangle$ & \quad\quad$a_8$: & $\langle\odot\neg s: a_1 \rangle$\\
\end{tabular}\smallskip

By Definition \ref{def_shattack} the shadow argument $a_8$ attacks $a_6$ as well as its sub-arguments $a_2 - a_5$. Moreover, it attacks $a_7$, which is a super-argument of $a_3$. As a result, none of the conclusions of arguments $a_2$-$a_7$ are \sys{DAF_{\odot}}-consequences of $\Gamma_{\ref{gsha}}$.
\end{example1}
Example \ref{ex_failmeta} no longer serves as a counter-example to properties \ref{prop_cautmon} and \ref{prop_ratmon} provided in Section \ref{sub_meta}. We can construct the shadow argument $a_{12}:\langle \odot s: a_5\rangle$.
By clause (i) of Definition \ref{def_shattack}, this argument attacks $a_4$ as well as its sub-arguments $a_1$-$a_3$. As a result of this attack, $\Gamma_{\ref{gcautmon}}\not\vdash_{\sys{DAF_{\odot}}}\OOs q$ and $\Gamma_{\ref{gcautmon}}\not\vdash_{\sys{DAF_{\odot}}}\OOs r$. More generally, we can show that the cautious monotonicity property (Property \ref{prop_cautmon} in Section \ref{sub_meta}) holds for \sys{DAF_{\odot}}. A proof is provided in Theorem \ref{thm:CM:shadow} of the Appendix.

Instead of -- and equivalently to -- working with the $\odot$-operator and Definitions \ref{def_doubtarg} and \ref{def_shattack}, we could have generalized Definitions \ref{def_fattack} and \ref{def_cattack} so as to include attacks on sub-arguments. Definitions \ref{def_fattack} and \ref{def_cattack} currently entail that if $a$ attacks $b$, then $a$ attacks all super-arguments of $b$. In the generalized form, these definitions would entail that if $a$ attacks $b$, then $a$ attacks all superarguments of all sub-arguments of $b$.

There are two additional reasons for working with the doubt operator $\odot$, however. First, this operator has a clear and intuitive meaning, and adds expressivity to our argumentation frameworks. Second, by characterizing shadow arguments via a separate operator we can think more transparently about (a) the implementation of additional logical properties of this operator, and (b) alternatives to Definition \ref{def_shattack}. Regarding (a), think about the strengthening rule (`If $\odot A$, then $\odot B$ whenever $B\vdash A$'), which carries some intuitive force. Regarding (b), reconsider Example \ref{ex_shadow}, and suppose we add the premise $\top\Ra\neg p$ to $\Gamma_{\ref{gsha}}$. A not-so-skeptical reasoner may argue that in this case we should not be able to cast doubt on the arguments $a_3$ and $a_7$, since the doubt casted on argument $a_4$ arguably arises in view of the conflicted conditional obligation to see to it that $p$.\footnote{Caminada's \emph{HY-arguments} from \cite{Cam04hy} are similar in spirit to this less skeptical proposal.}

\section{Related work}\label{sec_related}

Due to space limitations, we restrict our discussion of related formalisms to those of input/output logic (Section \ref{sub_iologic}) and those based on formal argumentation frameworks (Section \ref{sub_argfranor}). A comparison with other related deontic systems, such as Nute's defeasible deontic logic \cite{Nut97,Nut97b} and Horty's default-based deontic logic \cite{Hor03,Hor93,Hor97,Hor12} is left for an extended version of this article.

\subsection{Input/output logic}\label{sub_iologic}
Like the constrained input/output (I/O) logics from \cite{MakTor01}, the DAFs defined here are tools for detaching conditional obligations relative to a set of inputs and constraints. Unlike most I/O logics, none of these DAFs validates strengthening of the antecedent (SA) for conditional obligations -- from $A\Ra C$ to infer $(A\wedge B)\Ra C$. Unrestricted (SA) is counter-intuitive if we allow for conflict-resolution via logical analysis as defined Section \ref{sub_spec}, since it allows the unrestricted derivation of more specific from less specific conditional obligations.\footnote{In \cite{Sto10} an I/O system is presented which invalidates (SA) in the context of exempted permissions which are subject to conflict-resolution via logical analysis (specificity).}

\begin{example1}[\sys{DAF} and I/O logic]\label{sub_iodaf}
Let $\Gamma_{\Gtel\label{giodaf}} = \{p, p\Ra q, p\Ra\neg r, q\Ra r\}$. We construct the following arguments on the basis of $\Gamma_{\ref{giodaf}}$:\smallskip

\begin{tabular}{llll}
$a_1$: & $\langle \OOs q: p,p\Ra q \rangle$ & \quad\quad$a_3$: & $\langle \OOs r: a_1, q\Ra r \rangle$\\
$a_2$ & $\langle \OOs\neg r: p,p\Ra\neg r\rangle$ & \ & \ \\
\end{tabular}\smallskip

Since $a_2$ and $a_3$ attack each other in view of Definition \ref{def_cattack}, $a_2,a_3\not\in\mathcal{G}(\Gamma_{\ref{giodaf}})$, while $a_1\in\mathcal{G}(\Gamma_{\ref{giodaf}})$. Consequently, $\Gamma_{\ref{giodaf}}\not\vdash_{\sys{DAF}}\OOs r$ and $\Gamma_{\ref{giodaf}}\not\vdash_{\sys{DAF}}\OOs \neg r$ while $\Gamma_{\ref{giodaf}}\vdash_{\sys{DAF}}\OOs q$.
\end{example1}
In constrained I/O logic, triggered conditional obligations in the input are divided into maximally consistent subsets (MCSs). $\Gamma_{\ref{giodaf}}^{\Ra}$ has three MCSs: $\{p\Ra q, q\Ra\neg r\}$, $\{p\Ra q, p\Ra r\}$, and $\{q\Ra\neg r, p\Ra r\}$. In \cite{MakTor01} two ways are presented for dealing with conflicts and constraints: via a full meet operation on the generated MCSs, or via a full join operation on the generated MCSs. The first approach gives us none of $q,r$, and $\neg r$ for $\Gamma_{\ref{giodaf}}$. The second gives us all three.

Some of the I/O logics defined in e.g.\ \cite{MakTor00,MakTor01,ParTor14} validate intuitively appealing rules which are not generally valid in our DAFs, such as the rule (OR) -- from $A\Ra C$ and $B\Ra C$ to infer $(A\vee B)\Ra C$. A detailed study of the appeal and implementation of (OR) and similar rules in the present argumentative setting is left for future investigation.


\subsection{Formal argumentation}\label{sub_argfranor}

Several ways of modeling normative reasoning on the basis of formal argumentation have been proposed in the literature. For instance, the approach in \cite{gabbay2012bipolar} is based on bipolar abstract argumentation frameworks. Dung's abstract argumentation frameworks are enriched with a support relation that is defined over the set of abstract arguments. This device is used to express deontic conditionals. A similar idea is used in \cite{oren2008argumentation} where a relation for evidential support is introduced. Argumentation schemes of normative reasoning are there expressed by means of Prolog-like predicates and subsequently translated into an argumentation framework. Here, we follow the tradition of structured or instantiated argumentation in which no support relation between arguments is needed. In our approach conditional obligations are modeled by a dyadic operator \(\Rightarrow\) that is part of the object language. Arguments consist of sequences of applications of factual and deontic detachment. As a consequence, for instance, evidential or factual support is an intrinsic feature of our arguments and is modeled via the factual detachment rule.

The general setting of our DAFs is close to ASPIC$^+$. For instance, in the dynamic legal argumentation systems (in short, DLAS) from \cite{prakken2013formalising}, deontic conditionals are also modeled via a defeasible conditional \(\leadsto\) in the object language. There are several differences to our approach. For instance, our conditionals are not restricted to conjunctions of literals as antecedents. As a consequence we needed to define a strong fact attack rule (Def.~\ref{def_fattack}) that, in order to avoid contamination problems (see Ex.~\ref{ex_incoh}), warrants that arguments with inconsistent supports are defeated.\footnote{Other solutions to this problem have been proposed, e.g., in \cite{Wu-Phd}.} Our fact attack and our shadow attack rules do not conform to the standard attack types defined in ASPIC$^+$ (rebutting, undercutting, and undermining). Our conflict attacks can be seen as forms of ASPIC$^+$-type rebuttals where the contrary of \(\OOs A\) is defined by \(\OOs\neg A\).

Unlike DLAS or Horty's deontic default logics, we follow the tradition in deontic logic to have a dedicated operator \(\OOs\) for unconditional obligations which, for instance, allows to formally distinguish between cases of deontic and cases of factual detachment.

Recently, van der Torre \& Villata extended the DLAS approach with deontic modalities \cite{TorVil14}, adopting the input/output methodology from Section 4.1. The resulting systems, like \sys{DAF}, allow for versions of the factual and deontic detachment rules. Moreover, they allow for the representation of permissive norms. Unlike \sys{DAF}, and unlike the I/O logics from Section 4.1, these systems do not have inheritance (weakening) or aggregation rules.

Another approach in which formal argumentation is used for the analysis of traditional problems of deontic logic, such as contrary-to-duty and specificity cases is \cite{strasser2015normative}. There, arguments are Gentzen-type sequents in the language of standard deontic logic and conditionals are expressed using material implication. One drawback which is avoided in our setting is that there conditionals are contrapositable and subject to strengthening of the antecedent.



\section{Outlook}\label{sec_outlook}

We presented a basic logic, \sys{DAF}, for detaching conditional obligations based on Dung's grounded semantics for formal argumentation. We extended \sys{DAF} with mechanisms for conflict-resolution and for the anticipation of conflicts and violations. For now, these mechanisms mainly serve to illustrate the modularity of our framework. A detailed study of e.g.\ different approaches to prioritized reasoning, or different conceptions of specificity-based conflict-resolution, is left for an extended companion paper.

We conclude by mentioning three challenges for future research. The first is to include permission statements. The second is to increase the `logicality' of our framework by allowing for the nesting and for the truth-functional combination of formulas of the form $\OOs A,A\Ra B$, or $\Box A$. The third is to extend our focus beyond grounded extensions, and to study how our framework behaves when subjected to different types of acceptability semantics for formal argumentation. Working with Dung's preferred semantics \cite{Dun95}, for instance, allows for the derivation of so-called floating conclusions \cite{Hor02,MakSch91}.

\bibliographystyle{plain}
\bibliography{phdrefs}

\providecommand{\noopsort}[1]{}
\begin{thebibliography}{10}
\expandafter\ifx\csname url\endcsname\relax
  \def\url#1{\texttt{#1}}\fi
\expandafter\ifx\csname urlprefix\endcsname\relax\def\urlprefix{URL }\fi
\newcommand{\enquote}[1]{``#1''}

\bibitem{StrArg14}
Besnard, P., A.~Garcia, A.~Hunter, S.~Modgil, H.~Prakken, G.~Simari and
  F.~Toni~(eds.), \emph{Special issue: Tutorials on structured argumentation},
  Argument and Computation \textbf{5(1)} (2014).

\bibitem{Cam04hy}
Caminada, M., \emph{Dialogues and {HY}-arguments}, in: J.~Delgrande and
  T.~Schaub, editors, \emph{10th International Workshop on Non-Monotonic
  Reasoning (NMR 2004), Whistler, Canada, June 6-8, 2004, Proceedings}, 2004,
  pp. 94--99.

\bibitem{CamAmg07}
Caminada, M. and L.~Amgoud, \emph{On the evaluation of argumentation
  formalisms}, Artificial Intelligence \textbf{171} (2007), pp.~286 -- 310.

\bibitem{CarJon02}
Carmo, J. and A.~Jones, \emph{Deontic logic and contrary-to-duties}, in:
  D.~Gabbay and F.~Guenthner, editors, \emph{Handbook of Philosophical Logic
  (2nd edition) Vol.~8}, Kluwer Academic Publishers, 2002 pp. 265--343.

\bibitem{DelSch97}
Delgrande, J. and T.~Schaub, \emph{Compiling specificity into approaches to
  nonmonotonic reasoning}, Artificial Intelligence \textbf{90} (1997),
  pp.~301--348.

\bibitem{Dun95}
Dung, P., \emph{On the acceptability of arguments and its fundamental role in
  nonmonotonic reasoning, logic programming and n-person games}, Artificial
  Intelligence \textbf{77} (1995), pp.~321--357.

\bibitem{DunSon01}
Dung, P. and T.~Son, \emph{An argument-based approach to reasoning with
  specifity}, Artificial Intelligence \textbf{133} (2001), pp.~35--85.

\bibitem{gabbay2012bipolar}
Gabbay, D., \emph{Bipolar argumentation frames and contrary to duty
  obligations, preliminary report}, in: M.~Fisher, L.~van~der Torre, M.~Dastani
  and G.~Governatori, editors, \emph{Computational Logic in Multi-Agent
  Systems: Proceedings of the 13th International Workshop, CLIMA XIII,
  Montpellier, France}, Springer, 2012 pp. 1--24.

\bibitem{Han08}
Hansen, J., \emph{Prioritized conditional imperatives: problems and a new
  proposal}, Autonomous Agents and Multi-Agent Systems \textbf{17} (2008),
  pp.~11--35.

\bibitem{Hor93}
Horty, J., \emph{Deontic logic as founded on nonmonotonic logic}, Annals of
  Mathematics and Artificial Intelligence \textbf{9} (1993), pp.~69--91.

\bibitem{Hor97}
Horty, J., \emph{Nonmonotonic foundations for deontic logic}, in: D.~Nute,
  editor, \emph{Defeasible Deontic Logic: Essays in Nonmonotonic Normative
  Reasoning}, Kluwer Academic Publishers, 1997 pp. 17--44.

\bibitem{Hor02}
Horty, J., \emph{Skepticism and floating conclusions}, Artificial Intelligence
  \textbf{135} (2002), pp.~55--72.

\bibitem{Hor03}
Horty, J., \emph{Reasoning with moral conflicts}, No\^{u}s \textbf{37} (2003),
  pp.~557--605.

\bibitem{Hor07}
Horty, J., \emph{Defaults with priorities}, Journal of Philosophical Logic
  \textbf{36} (2007), pp.~367--413.

\bibitem{Hor12}
Horty, J., \enquote{{Reasons as Defaults},} Oxford University Press, 2012.

\bibitem{Mak94}
Makinson, D., \emph{General patterns in nonmonotonic reasoning}, in: D.~M.
  Gabbay, C.~J. Hogger and J.~A. Robinson, editors, \emph{Handbook of Logic in
  Artificial Intelligence and Logic Programming, Vol.~3}, Oxford University
  Press, 1994 pp. 35--110.

\bibitem{MakSch91}
Makinson, D. and K.~Schlechta, \emph{Floating conclusions and zombie paths: two
  deep difficulties in the ``directly skeptical'' approach to defeasible
  inheritance nets}, Artificial Intelligence \textbf{48} (1991), pp.~199--209.

\bibitem{MakTor00}
Makinson, D. and L.~van~der Torre, \emph{Input/output logics}, Journal of
  Philosophical Logic \textbf{29} (2000), pp.~383--408.

\bibitem{MakTor01}
Makinson, D. and L.~van~der Torre, \emph{Constraints for input/output logics},
  Journal of Philosophical Logic \textbf{30} (2001), pp.~155--185.

\bibitem{ModPra14}
Modgil, S. and H.~Prakken, \emph{The {ASPIC}+ framework for structured
  argumentation: a tutorial}, Argument \& Computation \textbf{5} (2014),
  pp.~31--62.

\bibitem{Nut97b}
Nute, D., \emph{Apparent obligation}, in: D.~Nute, editor, \emph{Defeasible
  Deontic Logic: Essays in Nonmonotonic Normative Reasoning}, Kluwer Academic
  Publishers, 1997 pp. 287--315.

\bibitem{Nut97}
Nute, D., editor, \enquote{Defeasible Deontic Logic: Essays in Nonmonotonic
  Normative Reasoning,} Kluwer Academic Publishers, 1997.

\bibitem{oren2008argumentation}
Oren, N., M.~Luck, S.~Miles and T.~Norman, \emph{An argumentation inspired
  heuristic for resolving normative conflict}, in: \emph{Proceedings of the
  fifth workshop on coordination, organizations, institutionsm and norms in
  agent systems, AAMAS-08, Toronto}, 2008, pp. 41--56.

\bibitem{ParTor14}
Parent, X. and L.~van~der Torre, \emph{``{S}ing and dance!'' {I}nput/output
  logics without weakening}, in: F.~Cariani, D.~Grossi, J.~Meheus and
  X.~Parent, editors, \emph{DEON (12th International Conference on Deontic
  Logic in Computer Science)},  Lecture Notes in Artificial Intelligence
  \textbf{8554} (2014), pp. 149--165.

\bibitem{Pra11}
Prakken, H., \emph{An abstract framework for argumentation with structured
  arguments}, Argument and Computation \textbf{1} (2011), pp.~93--124.

\bibitem{prakken2013formalising}
Prakken, H. and G.~Sartor, \emph{Formalising arguments about norms.}, in:
  \emph{JURIX}, 2013, pp. 121--130.

\bibitem{PraSer96}
Prakken, H. and M.~Sergot, \emph{Contrary-to-duty obligations}, Studia Logica
  \textbf{57} (1996), pp.~91--115.

\bibitem{Ros30}
Ross, D.~W., \enquote{{The Right and the Good},} Oxford University Press, 1930.

\bibitem{Sto10}
Stolpe, A., \emph{A theory of permission based on the notion of derogation},
  Journal of Applied Logic \textbf{8} (2010), pp.~97--113.

\bibitem{DBLP:journals/japll/Strasser11}
Stra{\ss}er, C., \emph{A deontic logic framework allowing for factual
  detachment}, Journal of Applied Logic \textbf{9} (2011), pp.~61--80.

\bibitem{strasser2015normative}
Stra{\ss}er, C. and O.~Arieli, \emph{Normative reasoning by sequent-based
  argumentation}, Journal of Logic and Computation  (in print),
  doi:10.1093/logcom/exv050.

\bibitem{TorVil14}
van~der Torre, L. and S.~Villata, \emph{An {ASPIC}-based legal argumentation
  framework for deontic reasoning}, in: S.~Parsons, N.~Oren, C.~Reed and
  F.~Cerutti, editors, \emph{Computational Models of Argument - Proceedings of
  {COMMA} 2014, Atholl Palace Hotel, Scottish Highlands, UK, September 9-12,
  2014}, 2014, pp. 421--432.

\bibitem{Wu-Phd}
Wu, Y., \enquote{Between Argument and Conclusion. Argument-based Approaches to
  Discussion, Inference and Uncertainty,} Ph.D. thesis, Universite Du
  Luxembourg (2012).

\end{thebibliography}

\appendix

\def\nc{\vdash_{\bf DAF}} \def\Gc{\mathcal{G}} \def\Ac{\mathcal{A}}

\noindent Suppose we have a DAF \(\langle \mathcal{A}(\Gamma), {\sf Att}(\Gamma) \rangle\) and  \(\Gc\) is the grounded extension of \(\langle \mathcal{A}(\Gamma), {\sf Att}(\Gamma) \rangle\). In the following we write \(\Delta \equiv \Delta'\) if \(\Delta\) and \(\Delta'\) are equivalent in \textbf{CL}, i.e., \(Cn_{\bf CL}(\Delta) = Cn_{\bf CL}(\Delta')\).

\begin{lemma1}
\label{lem:att:fact}
If \(a \in \Gc\) and \(b\) attacks \(a\) then there is a \(c\) that fact attacks \(b\).
\end{lemma1}
\begin{proof}
Suppose \(a \in \Gc\). Hence, there is a minimal \(i\) such that \(a \in \Gc_i\). Suppose \(b\) attacks \(a\). Hence, there is a \(a_{i-1} \in \Gc_{i-1}\) that attacks \(b\) in some subargument \(b_{i-1}\). If the attack is a conflict attack, \(b_{i-1}\) attacks \(a_{i-1}\). Since \(a_{i-1} \in \Gc_{i-1}\), there is a \(a_{i-2} \in \Gc_{i-2}\) that attacks \(b_{i-1}\) in some subargument \(b_{i-2}\). If the attack is a conflict attack we can find another \(a_{i-3} \in \Gc_{i-3}\) that attacks \(b_{i-2}\) in some subargument \(b_{i-3}\), etc. At some point we reach \(i-k = 0\). Note that \(a_{i-k}\) fact attacks \(b_{i-k}\) since otherwise \(b_{i-k}\) attacks \(a_{i-k}\) which  contradicts \(a_{i-k} \in \Gc_0\).
\end{proof}

\begin{lemma1}\label{lem:G:G1}
If \(a \in \Gc\) then \(a \in \Gc_{1}\).
\end{lemma1}
\begin{proof}
Follows immediately with Lemma \ref{lem:att:fact}.
\end{proof}
\def\UO{{\sf UO}}
\def\OOs{{\sf O}}

\begin{lemma1}
\label{lem:weakening}
If \(a \in \Gc_i\) and \(b\) is obtained from \(a\) by weakening then also \(b \in \Gc_i\).
\end{lemma1}
\begin{proof}
  Suppose \(b = \langle\OOs A': a \rangle\) is obtained from \(a = \langle\OOs A : \ldots\rangle\) by weakening via $\Box(A \supset A')$ and suppose \(c\) attacks \(b\). If \(c\) fact attacks \(b\) then it also fact attacks \(a\) since \(\langle\UO(a) \equiv \UO(b)\rangle\). If \(c = \langle\OOs{-}A' : \ldots\rangle\) conflict attacks \(b\) then we can obtain \(c' = \langle\OOs{-}A:c\rangle\) be weakening of \(c\). Since \(c'\) conflict attacks \(a\), \(a \in \Gc\) by Lemma \ref{lem:att:fact} \(c'\) is fact attacked by some \(d\). Since \(\UO(c) \equiv \UO(c')\) also \(c\) is fact attacked by \(d\).
\end{proof}

\begin{lemma1}
\label{fact:subarg:G}
Where \(a \in \Gc\) and \(b\) is a subargument of \(a\), also \(b \in \Gc\).
\end{lemma1}
\begin{proof}
  Suppose c attacks b. Since $a$ is a superargument of $b$, $c$ also attacks $a$ and thus $c$ is fact attacked by Lemma \ref{lem:att:fact}.
\end{proof}

\begin{definition1}
\label{def:aT}

Let  \(a = \langle \OOs A_1 : \ldots\rangle\) be an argument  with $\UO(a) = \{ A_1 ,\ldots, A_n\}$ and let \(\pi\) be a permutation of \(\lbrace 2, \ldots, n \rbrace\). We know that for each \(A_i \in \UO(a)\) there is a subargument \(b_i = \langle\OOs A_i : \ldots\rangle\) of \(a\). We construct the argument \(\overline{a}_{\pi} = \langle \OOs\bigwedge \UO(a) : \ldots\rangle\) as follows.
\begin{itemize}
\item \(a_1^{\pi} = \langle\OOs A_1 : \ldots\rangle = a = b_1\)
\item \(a_2^{\pi} = \langle\OOs(A_1 \wedge A_{\pi(2)}): a_1^{\pi}, b_{\pi(2)}\rangle\)
\item \(a_{i}^{\pi} = \langle\OOs((A_1 \wedge \ldots \wedge A_{\pi(i-1)}) \wedge A_{\pi(i)}): a_{i-1}^{\pi}, b_{\pi(i)}\rangle\)
\end{itemize}

When we write \(\overline{a}\) we refer to \(a_n^{\pi}\) for \(\pi = {\sf id}\) (i.e., \(\pi(i) = i\)).
\end{definition1}

The following fact follows in view of Definition \ref{def:aT} and the definition of $\UO$:
\begin{fact1}\label{fact:UO:equiv}
\(\UO(a) \equiv \UO(\overline{a})\)
\end{fact1}

\begin{lemma1}
\label{lem:ova}
Where \(a = \langle\OOs A_1 : \ldots\rangle\) and \(\UO(a) = \lbrace A_1,\ldots, A_n \rbrace\):
If \(a \in \Gc\) then also \(\overline{a}_{\pi} \in \Gc\) where \(\pi\) is an arbitrary permutation over \(\lbrace 2,\ldots,n \rbrace\).
\end{lemma1}
\begin{proof}
We show the claim by a parallel induction for all \(a_i^{\pi}\) where \(1 \le i \le n\).

\noindent \emph{Base case}: \(a_1^{\pi} = a\) and thus the claim holds by the supposition.

\noindent \emph{Induction step}: We show the claim holds for \(a_{i+1}^{\pi}\). Suppose \(c\) attacks \(a_{i+1}^{\pi}\).

\emph{Case 1}: \(c\) fact attacks \(a_{i+1}^{\pi}\). Then \(\Box\neg\bigwedge(\UO(a_i^{\pi}) \cup \UO(b_{\pi(i+1)})) \in Cn_{\bf L^{\Box}}(\Gamma)\). By the induction hypothesis, both \(a_i^{\pi}\) and \(\overline{b_{\pi(i+1)}}\) are in \(\Gc\). By weakening there is an argument \((\OOs\neg\bigwedge\UO(b_{\pi(i+1)}) : a_i^{\pi})\) which attacks \(\overline{b_{\pi(i+1)}}\). By Lemma \ref{lem:weakening}, this argument is also in \(\Gc\). This is a contradiction since then \(\Gc\) is not conflict-free.

\emph{Case 2}: \(c\) conflict attacks \(a_{i+1}^{\pi}\). Thus, \(c\) is of the form \(\langle\OOs\neg(A_1 \wedge A_{\pi(2)} \wedge \ldots \wedge A_{\pi(i+1)}) : \ldots\rangle\). By aggregation and weakening \(c' = \langle\OOs\neg A_{\pi(i+1)}: a_i^{\pi}, c\rangle \in \mathcal{A}(\Gamma)\). This argument attacks \(b_{\pi(i+1)}\). Since, by Lemma \ref{fact:subarg:G}, \(b_{\pi(i+1)} \in \Gc\), \(c'\) is fact attacked in view of Lemma \ref{lem:att:fact}. Hence \(\Box \neg \bigwedge (\UO(a_i^{\pi}) \cup \UO(c)) \in Cn_{\bf L^{\Box}}(\Gamma)\). By weakening, \(c'' = \langle\OOs\neg(A_1 \wedge A_{\pi(2)} \wedge \ldots \wedge A_{\pi(i)}), \overline{c}\rangle \in \mathcal{A}(\Gamma)\). This argument attacks \(a_i^{\pi}\) and since by the induction hypothesis, \(a_i^{\pi} \in \Gc\), and by Lemma \ref{lem:att:fact}, \(c''\) is fact attacked. Since in view of Fact \ref{fact:UO:equiv} \(\UO(c) \equiv \UO(c'')\), also \(c\) is fact attacked.
\end{proof}

\begin{lemma1}
\label{lem:closure:1}
Where \(a = \langle \OOs A : \ldots \rangle \in \Gc\) and \(b = \langle \OOs B : \ldots \rangle \in \Gc\), also the argument \(c = \langle \OOs(A \wedge B) : a, b \rangle\) obtained from \(a\) and \(b\) by aggregation is in \(\Gc\).
\end{lemma1}
\begin{proof}
Suppose some \(d\) attacks \(c\).

\emph{Case 1}: \(d\) fact attacks \(c\). Hence, \(\Box\neg\bigwedge(\UO(a) \cup \UO(b)) \in Cn_{\bf L^{\Box}}(\Gamma)\). Since by Lemma \ref{lem:ova} \(\overline{a} = \langle\OOs\bigwedge \UO(a) : \ldots\rangle \in \Gc(\Gamma)\), we also get by weakening and Lemma \ref{lem:weakening} that \(\langle\OOs\neg\bigwedge \UO(b) : \overline{a}\rangle \in \Gc(\Gamma)\). However, \(\overline{b} \in \Gc(\Gamma)\) by Lemma \ref{lem:ova}. This is a contradiction since \(\overline{a}\) attacks \(\overline{b}\).

\emph{Case 2}: \(d = \langle \OOs{-}D : \ldots \rangle\) conflict attacks \(c\). If \(D \in \UO(a)\) or \(D \in \UO(b)\), then \(d\) also conflict attacks \(a\) or \(b\). Thus, by Lemma \ref{lem:att:fact}, \(d\) is fact attacked. If \(d = \langle\OOs{-}(A \wedge B) : \ldots\rangle\) we construct the argument \(f = \langle\OOs(A \wedge {-}(A \wedge B)) : d, a\rangle\) and by weakening \(f' = \langle\OOs\neg B : f\rangle\). Since \(f'\) attacks \(b\) and by Lemma \ref{lem:att:fact}, \(f'\) is fact attacked. Hence, \(\Box\neg\bigwedge \UO(f') \in Cn_{\bf L^{\Box}}(\Gamma)\). Since \(\UO(f') = \UO(d) \cup \UO(a)\), also \(\Box\neg\bigwedge(\UO(d) \cup \UO(a)) \in Cn_{\bf L^{\Box}}(\Gamma)\). Hence, by weakening \(\overline{d}\) we get \(d' = \langle\OOs\neg\bigwedge \UO(a) : \overline{d}\rangle\). Since \(\overline{a} \in \Gc(\Gamma)\), \(d'\) attacks \(\overline{a}\) and by Lemma \ref{lem:att:fact}, \(d'\) is fact attacked. Since \(\UO(d') \equiv \UO(d)\) in view of Fact \ref{fact:UO:equiv}, also \(d\) is fact attacked.

Altogether, any attacker of \(c\) is fact attacked and hence \(c \in \Gc(\Gamma)\).
\end{proof}

The following lemma follows immediately.
\begin{lemma1}
\label{lem:closure:2}
If \(\Gamma \nc \OOs A\) and \(\Gamma \nc \OOs B\) then also \(\Gamma \nc \OOs(A \wedge B)\).
\end{lemma1}

\begin{theorem1}
\label{thm:closure}
If \(\Gamma \nc \OOs  A\) for each \(A \in \Delta\) and \(B \in Cn_{\bf CL}(\Delta)\), then \(\Gamma \nc \OOs B\).
\end{theorem1}
\begin{proof}
By compactness there is a finite \(\lbrace A_1, \ldots, A_n \rbrace \subseteq \Delta\) such that \(B \in Cn_{\bf CL}(\lbrace A_1, \ldots, A_n \rbrace)\). By multiple applications of Lemma \ref{lem:closure:2}, \(\Gamma \nc \bigwedge_{i=1}^n A_i\). By Lemma \ref{lem:weakening}, \(\Gamma \nc B\).
\end{proof}

\begin{theorem1}
\label{thm:consistency}
There is no \(A\) for which \(\Gamma \nc \OOs A\) and \(\Gamma \nc \OOs \neg A\).
\end{theorem1}
\begin{proof}
Assume for a contradiction that \(\Gamma \nc \OOs A\) and \(\Gamma \nc \OOs \neg A\). Thus, there are \(a = \langle \OOs A: \ldots \rangle \in \Gc(\Gamma)\) and \(b = \langle \OOs\neg A: \ldots \rangle \in \Gc(\Gamma)\). However, since \(b\) conflict attacks \(a\), this is not possible since \(\Gc(\Gamma)\) is conflict-free.
\end{proof}

\begin{theorem1}
\label{thm:CT}
Let \(\Delta_{\Rightarrow} = \lbrace \top \Rightarrow A \mid A \in \Delta \rbrace\). If \(\Gamma \nc \OOs A\) for all \(A \in \Delta\) and \(\Gamma \cup \Delta_{\Rightarrow} \nc \OOs B\), then \(\Gamma \nc \OOs B\).
\end{theorem1}
\begin{proof}
Let \(\Delta = \lbrace A_1, \ldots, A_n \rbrace\) and \(\Gamma' = \Gamma \cup \Delta_{\Rightarrow}\). By the supposition we know that for each \(i \in \lbrace 1, \ldots, n \rbrace\) there is a \(a_i = \langle \OOs A_{i} : \ldots\rangle \in \Gc(\Gamma)\). Let \(a_i^0 = \langle \OOs A_{i} : \top, \top \Rightarrow A_i \rangle \in \Ac(\Gamma')\). By the supposition, there is a \(b = \langle \OOs B : \ldots \rangle \in \Gc(\Gamma')\). Let \(b_{[a_1^0/a_1,\ldots, a_n^0/a_n]}\) be the result of replacing each subargument \(a_i^0\) by \(a_i\) in \(b\). Suppose now that some \(c\) attacks \(b_{[a_1^0/a_1,\ldots, a_n^0/a_n]}\) in \(\langle \Ac(\Gamma), {\sf Att}(\Gamma) \rangle\).

\emph{Case 1}: \(c\) also attacks \(b\) in \(\langle \Ac(\Gamma'), {\sf Att}(\Gamma') \rangle\). Hence, by Lemma \ref{lem:att:fact}, there is a \(e\) that fact attacks \(c\) and thus \(\Box\neg\bigwedge\UO(c) \in Cn_{\bf L^{\Box}}(\Gamma')\). Hence, also \(\Box\neg\bigwedge\UO(c) \in Cn_{\bf L^{\Box}}(\Gamma)\) and thus \(c\) is also fact attacked in \(\langle \Ac(\Gamma), {\sf Att}(\Gamma) \rangle\).

\emph{Case 2}: \(c\) does not attack \(b\) in \(\langle \Ac(\Gamma'), {\sf Att}(\Gamma') \rangle\). We have two cases: (a) \(c = \langle \OOs{-}C : \ldots \rangle\) for some \(C \in \bigcup_{i=1}^n \UO(a_i) \setminus \UO(b)\), or (b) \(c\) fact attacks \(b_{[a_1^0/a_1,\ldots, a_n^0/a_n]}\).

We first show that case (b) is not possible. In this case \(\Box\neg\bigwedge(\UO(a_1) \cup \ldots \cup \UO(a_n) \cup \UO(b)) \in Cn_{\bf L^{\Box}}(\Gamma)\). Hence, \(\Box(\bigwedge(\bigcup_{i=1}^n\UO(a_i)) \supset \neg \bigwedge \UO(b)) \in Cn_{\bf L^{\Box}}(\Gamma)\). Note that since \(a_i \in \Gc(\Gamma)\) for each \(i \in \lbrace 1, \ldots, n \rbrace\), by Lemma \ref{lem:ova}, also \(\overline{a_i} \in \Gc(\Gamma)\). Moreover, we know by multiple applications of Lemma \ref{lem:closure:2}, that also \(\hat{a} = \hat{a_n} \in \Gc(\Gamma)\) where \(\hat{a_2} = \langle \OOs \bigwedge(\UO(a_1) \cup \UO(a_2)) : \overline{a_1}, \overline{a_2} \rangle\), \ldots{}, \(\hat{a_n} = \langle \OOs \bigwedge_{i=1}^n \bigwedge\UO(a_{i}) : \hat{a_{n-1}}, \overline{a_n} \rangle\). By weakening and Lemma \ref{lem:weakening}, also \(\hat{a}' = \langle \OOs \neg\bigwedge\UO(b) : \hat{a} \rangle \in \Gc(\Gamma)\). Note that \(\hat{a}'\) attacks \(b\) in \(\langle \Ac(\Gamma'), {\sf Att}(\Gamma') \rangle\). Thus, by Lemma \ref{lem:att:fact}, \(\hat{a}'\) is fact attacked and \(\Box\neg\bigwedge\UO(\hat{a}') \in Cn_{\bf L^{\Box}}(\Gamma')\). Thus, also \(\Box\neg\bigwedge\UO(\hat{a}') \in Cn_{\bf L^{\Box}}(\Gamma)\). Hence, \(\hat{a}'\) is also fact attacked in \(\langle \Ac(\Gamma), {\sf Att}(\Gamma) \rangle\).  This contradicts \(\hat{a}' \in \Gc(\Gamma)\).

In case (a), \(c\) attacks some \(a_i\). Since \(a_i \in \Gc(\Gamma)\) and by Lemma \ref{lem:att:fact}, \(c\) is fact attacked in $\langle\Ac(\Gamma), {\sf Att}(\Gamma)\rangle$.

We have shown that every attacker of $b_{[a_1^0/a_1,\ldots, a_n^0/a_n]}$ is fact attacked in $\langle \Ac(\Gamma), {\sf Att}(\Gamma) \rangle$ and thus $b_{[a_1^0/a_1,\ldots, a_n^0/a_n]} \in \Gc(\Gamma)$.
\end{proof}

\def\UO{{\sf UO}}
We now move to DAFs with shadow attacks. In the following we will silently assume that Lemma \ref{lem:att:fact} also applies to argumentation frameworks with shadow attacks, but leave the simple proof to the reader.


\def\ncodot{\vdash_{\bf DAF_\odot}}
\begin{theorem1} \label{thm:CM:shadow}
Suppose \(\Gamma \ncodot \OOs A\) and \(\Gamma \ncodot \OOs B\) then \(\Gamma \cup \lbrace \top \Rightarrow A \rbrace \ncodot \OOs B\).
\end{theorem1}
\begin{proof}
  Let \(\Gamma' = \Gamma \cup \lbrace \top \Rightarrow A \rbrace\). Since $\Gamma \ncodot \OOs A$ and $\Gamma \ncodot \OOs B$, there are \(a = \langle \OOs A : \ldots \rangle \in \Gc(\Gamma)\) and \(b = \langle \OOs B : \ldots \rangle \in \Gc(\Gamma)\). Suppose \(c\) attacks \(b\) in \(\langle\Ac(\Gamma'), {\sf Att}(\Gamma')\rangle\). If \(c \in \mathcal{A}(\Gamma)\), it is fact attacked in \(\langle\Ac(\Gamma), {\sf Att}(\Gamma)\rangle\) and thus also in \(\langle\Ac(\Gamma'), {\sf Att}(\Gamma')\rangle\). Suppose now that \(c \in \mathcal{A}(\Gamma') \setminus \mathcal{A}(\Gamma)\). Thus, there is a subargument $\langle \OOs A : \top, \top \Rightarrow A \rangle$ of $c$.

\noindent\emph{Case 1:} \(c\) fact attacks \(b\). Then \(\Box\neg\bigwedge\UO(b) \in Cn_{\bf L^{\Box}}(\Gamma')\) and hence also \(\Box\neg\bigwedge\UO(b) \in Cn_{\bf L^{\Box}}(\Gamma)\). Thus, \(b\) is also fact attacked in \(\langle \Ac(\Gamma), {\sf Att}(\Gamma) \rangle\) which contradicts \(b \in \Gc(\Gamma)\).

\noindent\emph{Case 2:} \(c = (\OOs C : \ldots)\) conflict attacks \(b\). Let \(c_a \in \Ac(\Gamma)\) be the argument obtained by replacing the subargument $\langle \OOs A : \top, \top \Rightarrow A \rangle$ in $c$ by $a$. Let $c_{a}' \in \Ac(\Gamma)$ be a minimal subargument of $c_a$ with conclusion $\OOs C$. Since $c_a'$ attacks $b$, $b \in \Gc(\Gamma)$, and by Lemma \ref{lem:att:fact}, $c_a'$ is fact attacked by some $d = \langle \Box\neg\bigwedge\Theta : -- \rangle$ where $\Theta \subseteq \UO(c_a')$. Thus, there is also a $d_{\odot} = \langle \odot \bigwedge \Theta : d \rangle$ that shadow attacks $c_a'$. Define for any argument $g$, ${\rm base}(g)$ as the set of all subarguments based on factual detachment  $\langle \OOs G : K, K \Rightarrow G \rangle$ of $g$. Note that ${\rm base}(c_a') \subseteq {\rm base}(c_a) \subseteq {\rm base}(a) \cup {\rm base}(c)$. We distinguish two cases: (a) ${\rm base}(c_a') \cap {\rm base}(a) = \emptyset$ and (b) ${\rm base}(c_a') \cap {\rm base}(a) \neq \emptyset$. In case (a), ${\rm base}(c_a') \subseteq {\rm base}(c)$. Since $c_a'$ is fact attacked, also $c$ is fact attacked. In case (b), since $c_a'$ is shadow attacked by $d_{\odot}$ and minimal, also each subargument of $c_a'$ is shadow attacked by $d_{\odot}$. Since ${\rm base}(c_a') \cap {\rm base}(a) \neq \emptyset$, there is also a subargument $a'$ of $a$ that is shadow attacked by $d_{\odot}$. Since $a'$ cannot be defended from this attack and $a \in \Gc(\Gamma)$ this is a contradiction.

We have thus shown that $c$ is fact attacked. Since $c$ was arbitrary this is sufficient to show that $b \in \Gc(\Gamma')$.
\end{proof}

\end{document}